%% file: main.tex
\documentclass[10pt]{uq-report}

\def\SNL{Optimization and Uncertainty Quantification, Sandia National Laboratories, Albuquerque, NM, 87123}
\def\MICH{University of Michigan, 3053 FXB, 1320 Beal Avenue, Ann Arbor, MI. 48109, USA}

\include{preamble}

\usepackage{authblk}
\shortauthor{Gorodetsky, Jakeman, Geraci}
\corauthor{A. A. Gorodetsky}
\coremail{goroda@umich.edu}
\shorttitle{Multifidelity networks}
\funding{}

\author[1]{A. A. Gorodetsky}
\affil[1]{\MICH}
\author[2]{J.D. Jakeman}
\affil[2]{\SNL}
\author[2]{G. Geraci}
\title{MFNets: Data efficient all-at-once learning of multifidelity surrogates as directed networks of information sources}

\usepackage{xcolor}

\usepackage{tikz}
\usepackage{subcaption}
\usepackage{diagbox}
\usepackage{todonotes}

\usetikzlibrary{bayesnet}

\newcommand{\revv}[1]{#1}
\usepackage[normalem]{ulem} 

\begin{document}

\maketitle

\begin{abstract}
\revv{We present an approach for constructing a surrogate from ensembles of information sources of varying cost and accuracy. The multifidelity surrogate encodes connections between information sources as a directed acyclic graph, and is trained via gradient-based minimization of a nonlinear least squares objective. While the vast majority of state-of-the-art assumes hierarchical connections between information sources, our approach works with flexibly structured information sources that may not admit a strict hierarchy. The formulation has two advantages: (1) increased data efficiency due to parsimonious multifidelity networks that can be tailored to the application; and (2) no constraints on the training data -- we can combine noisy, non-nested evaluations of the information sources. Numerical examples ranging from synthetic to physics-based computational mechanics simulations indicate the error in our approach can be orders-of-magnitude smaller, particularly in the low-data regime, than single-fidelity and hierarchical multifidelity approaches.}
\end{abstract}

\include{body}

\bibliographystyle{abbrv}
\bibliography{references}

\end{document}

%% file: preamble.tex
\usepackage{graphicx} 
\usepackage{subcaption} 

\usepackage{natbib}

\usepackage{algorithm}
\usepackage{algorithmic}
\usepackage{hyperref}

\usepackage[textsize=tiny]{todonotes}

\usepackage{diagbox}
\usepackage{tikz}
\usetikzlibrary{matrix, arrows, positioning, shapes, decorations.markings}
\usetikzlibrary{patterns}
\tikzstyle{vecArrow} = [thick, decoration={markings,mark=at position
   1 with {\arrow[semithick]{open triangle 60}}},
   double distance=1.4pt, shorten >= 5.5pt,
   preaction = {decorate},
   postaction = {draw,line width=1.4pt, white,shorten >= 4.5pt}]
\pgfdeclarelayer{background}
\pgfdeclarelayer{foreground}
\pgfsetlayers{background,main,foreground}
\usetikzlibrary{positioning,calc,shapes.callouts,shapes.arrows,arrows,matrix}
\usetikzlibrary{bayesnet, positioning}

\tikzstyle{line} = [draw, -latex']

\tikzset{array/.style={circle,draw,ultra thick,fill=purple,minimum size=0.5cm,node distance=1.0,text=white,font=\footnotesize},}
\tikzset{darray/.style={circle,draw,fill=yellow,minimum size=0.5cm,node distance=1.0},} 
\tikzset{ind/.style={color=greenish}}
\tikzset{con/.style={color=redish},}

\tikzset{nodesurround/.style={fill=yellow!20,rounded corners, draw=black!50, dashed}}

\usepackage{mathrsfs}
\usepackage{amsmath, amssymb}
\usepackage{amsthm}

\usepackage{centernot}

\newcommand{\reals}{\mathbb{R}}
\newcommand{\integers}{\mathbb{Z}}
\newcommand{\posint}{\integers_+}

\newcommand{\diag}{\textrm{diag}}

\newcommand{\rev}[1]{{\color{black} #1}}

\newcommand{\nmodel}{M}
\newcommand{\nmodels}{\nmodel}

\newcommand{\xxspace}[1]{{\cal X}_{#1}}

\newcommand{\bvec}[1]{\boldsymbol{#1}}

\newcommand{\probd}{p}

\newcommand{\gpath}[1]{\normalfont{\textrm{path}}\left(#1\right)}
\newcommand{\roots}[1]{\normalfont{\textrm{roots}}\left(#1\right)}
\newcommand{\parents}[1]{\normalfont{\textrm{pa}}\left(#1\right)}
\newcommand{\child}[1]{\normalfont{\textrm{child}}\left(#1\right)}
\newcommand{\ancestors}[1]{\normalfont{\textrm{anc}}\left(#1\right)}

%% file: body.tex
\section{Introduction}
\label{sec:intro}
Exclusive use of data \revv{from} a single ``high-fidelity'' information source to make predictions of unseen outcomes of complex physical simulation and/or experiments is often computationally intractable due to the cost of obtaining data from the most accurate information sources. An increasingly important strategy to address this challenge is to fuse information from an ensemble of available sources of varying accuracy and cost into a single predictive model. In this paper, we propose a new \textit{multifidelity} surrogate framework for performing such fusion that improves the ability to make accurate predictions whenever only sparse numerical simulation and physical experiments data can be obtained.

Much of the multifidelity literature focuses on predicting statistics of a high-fidelity information source using Monte Carlo type sampling approaches. This outer-loop process requires sampling the distributions of the uncertain parameters and evaluating the information sources (\textit{i.e.} running either numerical or physical experiments) to compute statistics such as mean and variance. 
Multifidelity Monte Carlo methods reduce the classical Monte Carlo estimator variance, which is proportional to the ratio between the random variable variance and the number of samples, by introducing additional estimators that are correlated with the MC estimator~\cite{Lavenberg1978,Giles2008,Peherstorfer2016b,Haji2016,Geraci2017,Gorodetsky2020cv,de2020bi}. The resulting variance reduction is \revv{determined} by the magnitude of the correlation\revv{s}.

This paper focuses on multifidelity information fusion algorithms for constructing surrogates of high-fidelity data sources \revv{that} can be used for computing statistics \revv{and other} outer-loop processes such as optimization. Similarly to \rev{single-fidelity} surrogate methods~\cite{Rasmussen2006,Xiu2002,Marzouk2007}, multifidelity surrogate methods exploit smoothness to produce accurate approximations that converge quickly to the highest-fidelity function --- in some cases exponentially fast~\cite{Ng_E_AIAA_2012,Kennedy2000,Teckentrup_JWG_SIAMUQ_2015,HajiAli_NTT_CMAME_2016, Nachar2019}. The efficacy of using multifidelity (MF) surrogates was first identified in~\cite{Alexandrov_DLR_SO_1998}. This work used limited high-fidelity data to correct local low-fidelity approximations to reduce the cost of \rev{trust-region-based} optimization. Various adaptations of this discrepancy-modeling approach followed~\cite{Lewis2000,Eldred2004,Ng2012,Berchier2016} including multi-level~\cite{Teckentrup_JWG_SIAMUQ_2015} and multi-index~\cite{HajiAli_NTT_CMAME_2016}.

Discrepancy-based MF approaches often employ a form of {\it component-wise} optimization to construct the surrogate. Specifically, these methods use high-fidelity data to correct low-fidelity approximations which were built solely with low-fidelity data. An alternative strategy is to use an \revv{{\it all-at-once} approach which fuses data from all information sources to inform the approximations of all data sources. Such an all-at-once procedure was first proposed in \cite{Kennedy2000} to build a Gaussian process surrogate of an expensive simulation code by applying co-kriging to the noiseless output of multiple (two or more) correlated simulation codes. Furthermore~\cite{Le2014} extended this approach to efficiently make predictions via recursive co-kriging approaches in which the hyperparameters are simultaneously inferred. A number of similar methods based upon polynomial approximation have also been developed~\cite{Bryson_R_AST_2017,Rumpfkeil_B_AIAAJ_2020}. }

\revv{Regardless of the optimization strategy used, the} overwhelming majority of existing MF surrogate approaches \revv{presume a hierarchy of information sources, ordered by their predictive capability}. For example, \cite{Kennedy2000,Le2014,Narayan_GX_SISC_2014} effectively utilize a hierarchy of models --- typically \revv{trained via 
a component-wise
rather than an all-at-once procedure} --- of increasing fidelity to build surrogates that leverage models with increasing physics and/or numerical discretizations. This assumption can be too restrictive when it is difficult to order models based upon predictive utility per unit cost. \revv{Such a situation can occur} when there is a complex interplay between numerical errors and physical modeling. A small number of works have focused on developing methods for fusing information sources that do not admit a strict ordering of fidelity \cite{Liu_OCW_EAAI_2018,Lam_AW_AIAA_2015,Jakeman2020,HajiAli_NTT_CMAME_2016}. Each of these encode and exploit a specific relationship between models. Recently however, \cite{Gorodetsky_JGE_2019} developed a multi-information fusion framework (MFNets) that provides a general framework to encode and exploit prior knowledge regarding the relationships between data. Examples of prior knowledge include insight that two low-fidelity information sources are more closely aligned with the high-fidelity source in different regions of the parametric domain or that the magnitude of the discrepancy between QoI computed with successive finite element models decreases as the mesh is refined. 

The MFNets framework was primarily developed and analyzed in the context of sampling-based MF approaches. In this paper we extend these ideas to the context of surrogates. The MFNets framework uses prior knowledge to posit a network of latent variables to explain observed relationships between information sources; when building surrogates based upon linear subspace models, e.g. polynomial approximations, these latent variables correspond to the coefficients of the polynomial basis. Conditional independence relationships are then used to encode the \rev{prior knowledge} and produce compact representations of the joint density of all latent variables which enable efficient procedures for inferring the latent variables and thus building a multifidelity surrogate.

Whereas the original MFNets paper focused on relationships between the underlying \textit{parameters} of the approximation, in this paper we construct a multifidelity surrogate where the connections between information sources are focused on their \textit{observed outputs}. \revv{The method we present minimizes a standard least squares objective, motivated by (regularized) maximum-likelihood estimation that allows for noisy data and enables models of non-hierarchical, non-nested, and unstructured information sources.} The \rev{novel} contributions of this paper are the following
\revv{
\begin{enumerate}
\item \revv{Creation of a new modeling framework for constructing parsimonious multifidelity networks of surrogate models that are tailored to a given (possibly non-hierarchical) ensemble of information sources};
\item Development of a gradient-based, all-at-once, optimization procedure for learning algorithm for estimating the network weights; and
\item Numerical verification that the approach enables significant accuracy benefits over state-of-the-art hierarchical/recursive models.
\end{enumerate}
}

\revv{Finally we wish to remark that, although we are using a network of surrogates to fuse multiple information sources, the approach we present is significantly different to multifidelity methods based on neural networks (NNs)~\cite{Yan2020,Chakraborty2020,Meng2020}. To date, NN-based methods have only been applied to bi-fidelity model ensembles, with one high- and low-fidelity information source. Moreover, most of these methods construct surrogates in a component-wise fashion. In this paper we demonstrate that, for the problems tested, all-at-once non-hierarchical information fusion based upon networks of linear-subspace representations of each information source, significantly outperforms hierarchical multifidelity strategies. Our framework is general however, and could easily employ other approximation strategies, such as neural networks, for each information source within the multifidelity network. Moreover our approach improves interpretability and performance by directly associating training data with multiple layers in the network.}

The rest of this paper is structured as follows. Section~\ref{sec:mf_surrogate} introduces our new concept of multifidelity surrogate models. Section~\ref{sec:opt} describes the learning algorithm for estimating the multifidelity surrogate model parameters, and Section~\ref{sec:experiments} describes a large set of numerical experiments highlighting the applicability and benefits of our approach.

\section{Multifidelity surrogate models}\label{sec:mf_surrogate}
In this section we define the multifidelity surrogate model (MFnet). To this end, we first review the \revv{construction of single fidelity surrogate models using linear-subspace models}, and then we formally define the \revv{MFNets} surrogate as a network of such single-fidelity surrogates.  Using this definition, we then \revv{formulate} a nonlinear least squares regression problem\revv{, based upon maximum likelihood estimation,} for estimating the coefficients of the multifidelity approximation. Finally, we discuss some approximation properties of multifidelity networks.

\subsection{Notation}

Let $\posint$ denote the set of positive integers and $\reals$ the set of reals. Let $M \in \posint$ the number of information sources from which we obtain data in the form of input-output pairs $\left(x_k^{(j)}, y_k^{(j)}\right)_{j=1}^{n_k}$, where $n_{k} \in \posint$, $x_k^{(j)} \in \mathcal{X}_k \subseteq \reals^d$, $d \in \posint$, and $y_i^{(j)} \in \reals$. We will use bold letters to \revv{indicate} ordered collections of like items. \revv{For example the sets of training samples and associated values, respectively given by} $\bvec{x}_k = [x_k^{(1)},x_k^{(2)},\ldots, x_{k}^{(n_k)}]$ and $\bvec{y}_k = [y_k^{(1)},y_{k}^{(2)},\ldots,y_{k}^{(n_{k})}].$

\revv{Our goal is to learn the relationship between surrogates $f_k: \mathcal{X}_k \to \reals$ of each information source $k = 1,\ldots,M.$} When each surrogate $f_k$ is a linear subspace model, it is parameterized as a linear combination of functions $f_k(x;\theta) = V_k^T(x) \theta$ where $V:\mathcal{X}_k \to \reals^{p}$, $\theta \in \reals^p$ and $p \in \posint.$ We \revv{sometimes parameterize the} basis functions explicitly so that $V_k(x) = [v_{k1}(x),\ldots,v_{kp}(x)]$ for $v_{ki} : \xxspace{k} \to \reals$. If the bases $({v_{ki}})$ are complete in $L_2$ as $p \to \infty$, then this surrogate {\it converges for all functions in $L_2$.} When the bases are evaluated at $n$ inputs $\bvec{x}$, then $V^T_k(\bvec{x}) \in \reals^{n \times p}$ represents a \revv{Vandermonde-like matrix whose rows correspond to the basis functions evaluated at each input and whose columns correspond to the evaluation of a single basis function at all inputs.}

\revv{We use a directed acyclic graph to encode the relationship between the individual surrogates $f_{k}$, and this graph represents the multifidelity model.} A directed acyclic graph (DAG) $\mathcal{G}$ is a tuple $(\mathcal{V}, \mathcal{E})$ of nodes and edges, respectively, where the nodes are isomorphic to the positive integers and thus can be indexed $k=1,2,3,\ldots,M$. The graph consists of $M = |\mathcal{V}|$ nodes representing $M$ information sources. The directed edges $(j \to i)$ encode explicit dependencies between node (source) $j$ to node $i$. We refer to the parents of a node $k \in \mathcal{V}$ as those nodes that have an edge exiting them and entering $k$, i.e. $\parents{k} = \{ \ell \in \mathcal{V} : (\ell \to k) \in \mathcal{E}\}$. 
The children of a node are denoted by  $\child{k} = \{\ell \in \mathcal{V} : (k \to \ell) \in \mathcal{E}\}$. A path along the graph is a sequence of nodes along a set of directed edges. A path, denoted $\gpath{i_1,i_2,\ldots,i_m}$, exists on a DAG if $(i_j \to i_{j+1}) \in \mathcal{E}$ for $j=1,\ldots,m-1$. Finally, we will denote the ancestors of a node $k$ by $\ancestors{k}$. The ancestors are all those nodes $\ell$ from which there exists a path in $\mathcal{G}$ to node $k$. The \textit{roots} of the graph are those nodes with no parents.

\subsection{Single fidelity surrogates}

\revv{In this paper we minimize a nonlinear least squares objective, derived using standard maximum likelihood arguments,  to train a multifidelity network of surrogates. To facilitate this discussion, we first review the derivation of the standard least squares problem for training single-fidelity approximations. We then extend this procedure to the multifidelity setting.} Under the assumption that the observations are corrupted by independent Gaussian noise with zero mean and variance $\sigma^2$, the likelihood of observing the data with the linear subspace model is 
\begin{equation*}
  \probd(y_k \mid x, \theta) = (2\pi\sigma^2)^{-1/2}\exp\left(-\frac{1}{2\sigma^2}\left(y_k -V_k^T(x)\theta\right)^2\right)
\end{equation*}
If we obtain $n$ independent data points, then \rev{the} likelihood of the ensemble is
\begin{equation}\label{eq:sf-liklihood}
  \probd(y_k^{(1)},\ldots,y_k^{(n)} \mid x^{(1)},\ldots, x^{(n)}, \theta) = (2\pi\sigma^2)^{-n/2}\prod_{i=1}^{n}\exp\left(-\frac{1}{2\sigma^2} \left(y_k^{(i)} -V_k^T(x^{(i)})\theta\right)^2\right).
\end{equation}
We can then estimate the parameters $\theta$ by maximizing the log of the likelihood, with the following optimization problem
\begin{equation}\label{eq:linear-lstsq}
  \theta^{*} = \arg \min \sum_{i=1}^n\left(y_k^{(i)} -V_k^T(x^{(i)})\theta\right)^2  = \revv{\left(V_k(\bvec{x}) V_k^T(\bvec{x})\right)^{-1} V_k(\bvec{x}) \bvec{y}_k,}
\end{equation}
\revv{which is a linear least-squares regression objective with the closed form solution given in \eqref{eq:linear-lstsq}.}

\subsection{Multifidelity network surrogates}
\revv{In this section we define a multifidelity surrogate that holistically models a network of single-fidelity models as a DAG.}
\begin{definition}[Multifidelity surrogate]\label{def:mf_surr}
  A multifidelity surrogate is directed acyclic graph $\mathcal{G} = \left(\mathcal{V}, \mathcal{E}\right)$ with nodes corresponding to functions $\mathcal{V} = \{f_1,\ldots, f_{\nmodels}\}$ and directed edges $\mathcal{E} = \{(j \to i)\}$ representing connections between a function and its parents according to
  \begin{equation}
    f_{i}(x) = \sum_{j \in \parents{i}} \rho_{ji}(x)f_{j}(x)  + \delta_{i}(x). \label{eq:mf_surr}
  \end{equation}
  The root functions are represented by
  \begin{equation}
    f_{i}(x) = \delta_i(x).
  \end{equation}
  The edges and nodes are parameterized by linear-subspace models for the \revv{weighting} functions $\rho_{ji}$ and \revv{bias} functions $\delta_i$
  \begin{equation}
    \rho_{ji}(x) = W_{ji}^T(x) \alpha_{ji} \quad \textrm{ and } \quad \delta_{i}(x) = V_{i}^T(x) \beta_{i},
  \end{equation}
  respectively.
  The high-fidelity model is represented by a \revv{leaf node.}
\end{definition}

Simply, Definition \ref{def:mf_surr} states that given noisy data $(x,y)$ about some information source $k$, then the map $f_k$ from the features $x$ to the values $y$ is written as a  \rev{spatially-dependent} combination of a subset of other models $f_{j}$ and a discrepancy $\delta_{k}$. Specific cases of this approach have been considered previously. For example, hierarchical multifidelity methods, e.g. \cite{Kennedy2000}, assume that
\begin{equation}
  f_{k}(x) = \rho_{k-1,k}(x) f_{k-1}(x) + \delta_{k}(x).
\end{equation}

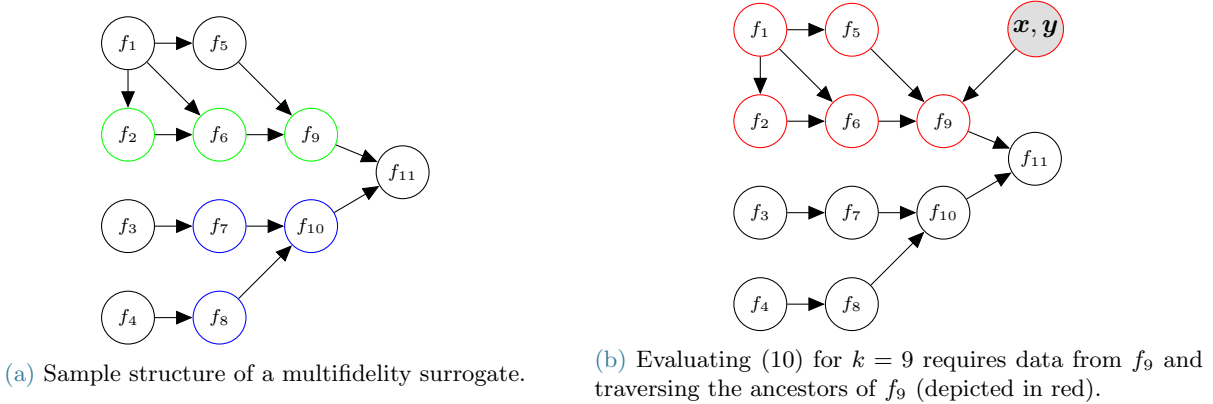
\begin{figure}
  \centering
  \tikzstyle{func} = [latent,  node distance=0.5cm, font=\scriptsize]
  \begin{subfigure}{0.49\textwidth}
    \centering
  \begin{tikzpicture}
    \node[func, ] at (0,0) (M1) {$f_{1}$} ;
    \node[func, below=of M1, draw=green] (M2) {$f_{2}$};
    \node[func, below=of M2] (M3) {$f_3$};
    \node[func, below=of M3] (M4) {$f_4$};

    \node[func,  right=of M1] (M5) {$f_5$} ;
    \node[func, below=of M5, draw=green] (M6) {$f_6$};
    \node[func, below=of M6, draw=blue] (M7) {$f_7$};
    \node[func, below=of M7, draw=blue] (M8) {$f_8$};

    \node[func, right=of M6, draw=green] (M9) {$f_9$};
    \node[func, right=of M7, draw=blue] (M10) {$f_{10}$};
    
    \node[func, right=of M9, yshift=-0.5cm] (M11) {$f_{11}$};

    \edge {M1} {M5};
    \edge {M1} {M6};
    \edge {M1} {M2};

    \edge {M2} {M6};
    \edge {M3} {M7};
    \edge {M4} {M8}; 
    
    \edge {M5} {M9};
    \edge {M6} {M9};
    
    \edge {M7} {M10};
    \edge {M8} {M10};

    \edge {M9} {M11};
    \edge {M10} {M11};
  \end{tikzpicture}
  \caption{Sample structure of a multifidelity surrogate. }
  \label{fig:dag1}
  \end{subfigure}
  ~
  \begin{subfigure}{0.49\textwidth}
    \centering
  \begin{tikzpicture}
    \node[func, draw=red] at (0,0) (M1) {$f_{1}$} ;
    \node[func, below=of M1,draw=red] (M2) {$f_{2}$};
    \node[func, below=of M2] (M3) {$f_3$};
    \node[func, below=of M3] (M4) {$f_4$};

    \node[func, right=of M1, draw=red] (M5) {$f_5$} ;
    \node[func, below=of M5, draw=red] (M6) {$f_6$};
    \node[func, below=of M6] (M7) {$f_7$};
    \node[func, below=of M7] (M8) {$f_8$};

    \node[func, right=of M6, draw=red] (M9) {$f_9$};
    \node[func, right=of M7] (M10) {$f_{10}$};
    
    \node[func, right=of M9, yshift=-0.5cm] (M11) {$f_{11}$};

    \node[obs,draw=red, above right=of M9] (data) {$\bvec{x}, \bvec{y}$}; 

    \edge {data} {M9};
    
    \edge {M1} {M5};
    \edge {M1} {M6};
    \edge {M1} {M2};

    \edge {M2} {M6};
    \edge {M3} {M7};
    \edge {M4} {M8}; 
    
    \edge {M5} {M9};
    \edge {M6} {M9};
    
    \edge {M7} {M10};
    \edge {M8} {M10};

    \edge {M9} {M11};
    \edge {M10} {M11};
  \end{tikzpicture}
  \caption{Evaluating \eqref{eq:likelihood-k} for $k=9$  requires data from $f_9$ and traversing the ancestors of $f_9$ (depicted in red).}
  \label{fig:sample_data}
  \end{subfigure}
  \caption{An example DAG used to define a multifidelity surrogate. This structure exhibits a complicated relationship between each function and the high-fidelity $f_{11}$. Both hierarchical and peer relationships are exhibited within these networks. For instance the left panel shows an example of hierarchical structure $(f_2 \to f_6 \to f_9)$ in green and example of peer structure $(f_7 \to f_{10}, f_{8} \to f_{10})$ in blue.}
  \label{fig:mfsurr}
\end{figure}

Our more general multifidelity surrogate formulation is the functional-space equivalent to the network-modeling strategy we introduced in~\cite{Gorodetsky_JGE_2019}, and therefore will be called an MFNet as well.  \rev{An} example MFNet is shown in Figure~\ref{fig:dag1}. Two important structures for multifidelity modeling are highlighted. The green nodes represent a hierarchical structure connecting information sources that can be ordered clearly according to predictive utility per unit cost. The blue nodes represent a peer structure that connects two low-fidelity sources\revv{, $f_7$ and $f_8$,} which \revv{may not be ordered by fidelity}, with a \revv{higher-fidelity source ($f_{10}$)}.

\begin{remark} The model in \eqref{eq:mf_surr} assumes a linear relationship between the pointwise evaluation of an information source and any of its ancestors. In the context of model discrepancy, this refers to both additive and scaling ``model error'' considerations and is commonly done in the literature~\cite{Kennedy2000}. Nevertheless, nonlinear approximations, such as those in~\cite{Perdikaris2017} are equally plausible in this work. For example we could use a nonlinear activation function \revv{$a(\cdot)$} to obtain
  \begin{equation*}
    f_{i}(x) = a\left(\sum_{j \in \parents{i}} \rho_{ji}(x)f_{j}(x)\right)  + \delta_{i}(x),
  \end{equation*}
  or
  \begin{equation*}
    f_{i}(x) = a\left(\sum_{j \in \parents{i}} \rho_{ji}(x)f_{j}(x)  + \delta_{i}(x)\right).
  \end{equation*}
  However, our aim is to demonstrate that there is a rich extension to the predominant approaches based on hierarchical/recursive modeling. Introducing sparse sets of parents for each of the information sources allows us to address a more complicated set of multifidelity relationships than exists in the literature, while simultaneously retaining a simple and data-efficient model to learn. However, all of the subsequent algorithmic work can be extended to more complicated relationships.
\end{remark}

\subsection{Multifidelity likelihood model}
In this section we derive an optimization objective that can be used to train \revv{all parameters associated with the multifidelity surrogate of the network at once.} Given a fixed graph, this procedure is responsible for fitting the $\alpha_{ji}$ and $\beta_{i}$, \revv{introduced in Definition~\ref{def:mf_surr}, associated with all nodes and edges in that DAG}. %

\revv{The training objective function is derived through the specification of likelihoods. We first consider the likelihood of information source $k$ and then obtain the final objective by combining the log likelihoods for each source}.
We make the standard assumption that the data for each node/model is corrupted by a zero-mean Gaussian \revv{error} with standard deviation $\sigma_{k}$. When $k$ is a root note, it has no ancestors and the likelihood is exactly the same as the single-fidelity likelihood \eqref{eq:sf-liklihood}
\begin{equation}
  \probd( \bvec{y}_k \mid \bvec{x}, \mathcal{G}) = \left(\frac{1}{\sqrt{2\pi}\sigma_k}\right)^{n}\exp\left(-\frac{1}{2\sigma_{k}^2} \sum_{i=1}^n \left(y_k^{(i)} - V_k^T\left(x^{(i)}\right) \beta_k \right)^2 \right),
\end{equation}
with corresponding negative log likelihood
\begin{equation}
  -LL_k(\beta_k) = \frac{n}{2}\log 2\pi + \revv{n} \log \sigma_k  + \frac{1}{2\sigma_k^2} \left \lVert \bvec{y}_k - V_k^T\left(\bvec{x}\right) \beta_k \right \rVert_{\revv{2}}^2.
\end{equation}
If node $k$ is not a root node, then the likelihood is
\begin{equation}
  \probd(\bvec{y}_k \mid \bvec{x}, \mathcal{G}) = \left(\frac{1}{\sqrt{2\pi}\sigma_k}\right)^{n}\exp\left(-\frac{1}{2\sigma_{k}^2} \sum_{i=1}^n \left(y_k^{(i)} - \left(\sum_{j \in \parents{k}} f_{j}(x^{(i)}; \gamma_j) W_{jk}^T(x^{(i)})\alpha_{jk}  + V_k^T\left(x^{(i)}\right) \beta_k \right)\right)^2 \right),
\end{equation}
with corresponding negative log likelihood, written as a function of only the relevant graph parameters, 
\begin{align}
  -LL_k\left(\beta_{k}, \{\alpha_{jk}, \gamma_j ; j \in \parents{k} \}\right) &=  \frac{n}{2}\log 2\pi + n \log \sigma_k  + \nonumber \\
  &\qquad \frac{1}{2\sigma_k^2} \sum_{i=1}^n \left(y_k^{(i)} - \left(\sum_{j \in \parents{k}} f_{j}(x^{(i)}; \gamma_j) W_{jk}^T(x^{(i)})\alpha_{jk}  + V_k^T\left(x^{(i)}\right) \beta_k \right)\right)^2,
  \label{eq:likelihood-k}
\end{align}
where $\gamma_j = \{ \alpha_{\ell i} : \ell,i \in \ancestors{j}\} \cup \{\beta_\ell : \ell \in \ancestors{j}\}$ denotes the set of parameters of node $j$ and its ancestors.

The likelihood of observing all data given the graph is simply the product \revv{$\prod_{k\in \mathcal{V}}^M \probd(\bvec{y}_k \mid \bvec{x}, \mathcal{G})$} so that the total negative log-likelihood becomes
\begin{align}\label{eq:graph-neg-log-like}
-LL(\mathcal{G}) =-\sum_{k\in \mathcal{V}}^M LL_k\left(\beta_{k},\{\alpha_{jk}, \gamma_j ; j \in \parents{k}\}\right).
\end{align}

This likelihood can be evaluated efficiently by recursing over the graph, starting with the highest-fidelity data. This recursion is efficient because evaluation of \eqref{eq:likelihood-k} for the $k$th node only requires traversing the ancestors of $k$ in the graph. For example in Figure~\ref{fig:sample_data} evaluating \eqref{eq:likelihood-k} for $k=9$ only requires visiting that node and its ancestors $k=1,2,5,6$. Because of the products between parents and edge parameters $\alpha_{jk}$, this objective results in a nonlinear least-squares problem. We will outline a gradient-based optimization procedure that leverages the graph structure for fast computation in Section~\ref{sec:opt}.

\subsection{Priors and regularization}

In some situations learning can be improved by using regularizing priors on the surrogate parameters. Here, we consider adding priors to the parameters of each edge function $\rho_{ij}(x; \alpha_{ij})$ and each node function $\delta_{i}(x; \beta_{k})$. If the priors are in the exponential family, then their logs can be easily added to the negative log likelihood~\eqref{eq:graph-neg-log-like} to \rev{obtain} a regularized learning problem.

\revv{In the absence of} additional information, we have assumed that the parameters are independent. As a result, the prior factorizes as
\begin{equation}
  \probd(\{\beta_{i} : i \in \mathcal{V}\}, \{\alpha_{ij}: i\to j \in \mathcal{E}\}) = \prod_{i \in \mathcal{V}} \probd(\beta_{i}) \prod_{j \in \parents{i}} \probd(\alpha_{ji}).
\end{equation}
We now assume that each of these parameters is in the exponential family and takes the form $\probd(\theta) = g(\theta)\exp(\phi^T(\theta) \nu)$ for some scalar valued functions $g(\theta)$, vector-valued function $\phi$, vector $\nu$ --- all of appropriate sizes. Then, taking the log of the prior we can obtain the following regularized optimization problem
\begin{equation}
  \text{minimize} -LL(\mathcal{G}) + \sum_{i \in V} \left(\log g(\beta_i) + \phi^T(\beta_i) \nu + \sum_{j \in \parents{i}}  \left(\log g(\alpha_{ji}) + \phi^T(\alpha_{ji}) \nu\right)\right),
  \label{eq:regularized_learning}
\end{equation}
where for simplicity we have assumed that all of the prior distributions are \rev{from} the same family. For Gaussian priors we obtain
\begin{equation}
  \text{minimize} -LL(\mathcal{G}) +  \sum_{i \in V} \left(\lambda_{i} \lVert \beta_i \rVert_2^2 + \sum_{j \in \parents{i}} \lambda_{ij}\lVert \alpha_{ji} \rVert_2^2 \right),
  \label{eq:gauss_prior}
\end{equation}
and for Laplace priors, which we utilize in Section~\ref{sec:dfat} to encourage sparsity, we obtain
\begin{equation}
  \text{minimize} -LL(\mathcal{G}) +   \sum_{i \in V} \left(\lambda_{i}\lVert \beta_i \rVert_1 + \sum_{j \in \parents{i}} \lambda_{ij}\Vert \alpha_{ji} \rVert_1 \right).
  \label{eq:laplace_prior}
\end{equation}
In the above, $\lambda_{ij}$ represents a \revv{penalty} that balances the regularization and likelihood terms.

\subsection{Discussion and relationship with hierarchical approaches}
In this section we comment on the approximation capacity of MFNet surrogates. While the specific approximation quality of a network will depend on the relationships amongst the functions, there are a couple of general comments that can be made. For this discussion let us assume that all $V_i(x)$  and $W_{ij}(x)$ \rev{consist} of polynomials up to order $p$. So that within a given setting $\rho_{ij}(x)$ and $\beta_{j}(x)$ are both multivariate polynomials of order $p$.
Now consider a function $k$ which is the weighted sum of $|\parents{k}|$ polynomials corresponding to the \revv{weighting} factors and a single $p$ order polynomial for the \revv{bias}
\begin{equation}
  f_{k}(x) = \sum_{\ell \in \parents{k}} f_{\ell}(x) W_{\ell k}^T(x) \alpha_{\ell k} + V_{k}^T(x) \beta_{\ell}.
\end{equation}
Since the roots of the graph are also polynomials of order $p$, all their descendants must be polynomials of greater order. Specifically, if each $f_{\ell}(x)$ is of polynomial order $m$, then the total polynomial order of its child $f_{k}$ is $m+p$. By induction, the order of a polynomial $f_{k}$ is then $hp$ where $h$ is the longest chain that leads to $k$. This induction argument proves the following proposition.
\begin{proposition}[MFNet approximation order of the ``highest-fidelity'' function.] \label{prop:high_fidelity}
Consider a \revv{weakly-connected}\footnote{\revv{A directed graph is weakly connected if the graph obtained by replacing the directed edges with undirected ones is connected. In other words, there is a path between every pair of nodes in the graph, if direction of the edge is ignored.}} MFNet $\mathcal{G} = (\mathcal{V},\mathcal{E})$, where $\rho_{ij}(x)$ and $\delta_{i}(x)$ are at most order $p$, for $i,j \in \mathcal{V}$. Let $f_k$ correspond to the ``high-fidelity'' model --- the one for which $\ancestors{k} \cup \{k\} = V.$ If the \rev{maximum-length} path from a root node to $f_k$ has $h$ nodes, then $f_k$ is a polynomial of order $hp.$
\end{proposition}
This proposition can also be used for any subgraph of $\mathcal{G}$ to determine the order of the surrogate at any fidelity. While this result suggests that all one needs to consider is a graph of the longest path, there can be advantages to using \revv{shallower network arrangements if one can exploit non-hierarchical relationships that exist in the ``true'' data generation process}. Next, we describe these advantages by means of an example.

Let us consider peer ($\mathcal{E} = \{(1 \to 3), (2 \to 3) \}$) and hierarchical ($\mathcal{E} = \{(1 \to 2), (2 \to 3)\})$  networks for a three model surrogate and assume that the lowest fidelity model $f_1$ is so inexpensive that we can obtain enough data to learn it exactly. Now suppose the \revv{true generative model for the data} is the peer graph, and our goal is to recover $f_3$. We will show that it is both simpler to optimize, and more data efficient to learn, the peer model rather than the hierarchical network --- even though both networks can represent the high-fidelity function $f_3$ easily. 

For the peer graph we have
\begin{equation}\label{eq:peer-motivating-example}
  f_3(x) = \begin{bmatrix}
    f_1(x) & 1
  \end{bmatrix}
  \begin{bmatrix}
    \rho_{13}(x; \alpha_{13}) \\
    \rho_{23}(x; \alpha_{23})\delta_{2}(x; \beta_{2}) + \delta_{3}(x; \beta_{3})
  \end{bmatrix}
    =
  \begin{bmatrix}
    f_1(x) & 1
  \end{bmatrix}
  \begin{bmatrix}
    \hat{\rho}_{13}(x; \alpha_{13}) \\
    \hat{\delta}_{3}(x; \beta_{2}, \beta_{3}, \alpha_{23})
  \end{bmatrix}.
\end{equation}
Now if all the functions \revv{$\rho_{ij}, \delta_i$ are total-degree polynomials} order $p$
\revv{, the final representation of $f_3$ is of order $2p$.}
Compare this setup with the hierarchical graph
\begin{equation}\label{eq:hier-motivating-example}
  f_3(x) =
  \begin{bmatrix}
    f_1(x) & 1
  \end{bmatrix}
  \begin{bmatrix}
    \rho^h_{12}(x; \alpha^h_{12})\rho^h_{23}(x; \alpha^h_{23}) \\
    \delta^h_{2}(x; \beta^h_{2})\rho^h_{23}(x; \alpha^h_{23}) + \delta^h_{3}(x; \beta^h_{3})
  \end{bmatrix}
  =\revv{
  \begin{bmatrix}
    f_1(x) & 1
  \end{bmatrix}
  \begin{bmatrix}
    \hat{\rho}^{h}_{13}(x; \alpha^{h}_{12}, \alpha^{h}_{23}) \\
    \hat{\delta}^h_{3}(x; \beta^h_{2}, \beta^h_{3}, \alpha^{h}_{23})
  \end{bmatrix},}
\end{equation}
 \revv{which also represents $f_3$ with a polynomial of degree $2p$; here the superscripts with $h$ distinguish the functions and parameters from the peer case. If the peer model \eqref{eq:peer-motivating-example} is the true generating process then, then setting $\hat{\rho}_{13}(x; \alpha_{13})=\hat{\rho}^{h}_{13}(x; \alpha^{h}_{12}, \alpha^{h}_{23})$ makes it evident that the hierarchical model~\eqref{eq:hier-motivating-example} is over-parameterized. The hierarchical function $\hat{\rho}^{h}_{13}$ has degree $2p$ but the peer function $\hat{\rho}_{13}$ is only degree $p$. The hierarchical multifidelity surrogate thus possesses an additional ${d+2p\choose d}-{d+p\choose d}$ unknown coefficients. The number of these extra coefficients grows quickly, with degree $p$ and dimension $d$, and consequently can make data requirements for learning also grow rapidly.}

Next, suppose that the hierarchical approach discards any knowledge of the low-fidelity function $f_1$. This approach, while \rev{counter-intuitive}, may be advantageous since it will avoid the need to recover \revv{the additional coefficients} $\alpha_{12}$. In this case, we have $f_2(x) = \delta^{\revv{rr}}_{2}(x)$ so that 
\begin{equation}
  f_{3}(x) = \hat{\rho}_{23}^{rr}(x) \hat{\delta}^{rr}_{2}(x) + \delta^{\revv{rr}}_{3}(x)
\end{equation}
where again the unknown forms require identifying an order $2p$ function ($\hat{\rho}_{23}^{rr} \hat{\delta}^{rr}_{2}$); \revv{here the superscript $rr$ serves to distinguish these approximations from the corresponding approximations above when the first model was not ignored.} Thus, discarding the $f_1$ data does create an easier problem for the hierarchical network \rev{from the perspective of reducing the number of unknowns, but is unable to leverage the $f_1$ information and therefore will have to compensate for this missing connection with potentially more complicated edge \revv{$\hat{\rho}_{23}^{rr}(x)$} and node functions \revv{$\delta^{rr}_{3}$} than necessary.}

\revv{Finally we remark that if the true data generating process is the hierarchical model then the peer model will be less efficient. Similarly, both these networks will be inefficient if a fully connected network generates the data. The goal of this paper is not to motivate peer networks but rather to show that our more general framework can represent a greater variety of problem cases.}

\section{Learning algorithm}\label{sec:opt}
In this section we describe how to leverage the graph structure to simultaneously compute the value and the gradient of the negative log likelihood for use within an optimization scheme. We derive the derivative with respect to all the graph parameters, and show that it can be reformulated as an efficient forward-backward sweep across the graph. The forward sweep evaluates all the ancestors and the backward sweep updates the gradients of the parameters. This procedure is essentially analogous to those used by software frameworks where computations are defined by a computational \rev{graph}, such as TensorFlow and PyTorch; however, we specialize it for the specific structure and relationships given here.

\subsection{Recursive structure of the gradient}
If $k$ is a root node, the likelihood is quadratic in $\beta$ so the gradient is 
\begin{equation*}
  \frac{ \partial(-LL_k)}{\partial \beta k} = -\frac{1}{2\sigma_k^2}\bvec{r}_k^TV_k^T(\bvec{x}), \label{eq:root_deriv}
\end{equation*}
where $\bvec{r}_k = \bvec{y}_k - V_k^T\left(\bvec{x}\right) \beta_k$ is the residual.

For non-root nodes $k$, we have to compute the gradient with respect to all parameters of the subgraph of ancestors. Let denote the residual between the data and approximation of the $k$th model as 
\begin{equation*}
  \bvec{r}_k = \bvec{y}_k - \left(\sum_{j \in \parents{k}} \diag(f_{j}(\bvec{x}; \gamma_j)) W_{jk}^T(\bvec{x})\alpha_{jk}  + V_k^T\left(\bvec{x}\right) \beta_k \right)
\end{equation*}
then the gradient with respect to $\alpha_{jk}$ is
\begin{equation*}
  \frac{\partial (-LL_k)}{\partial \alpha_{jk}} = -\frac{1}{\sigma_{k}} \bvec{r}_k^T f_{j}(\bvec{x}; \gamma_j) W_{jk}^T(\bvec{x}), \quad  \forall j \in \parents{k}.
\end{equation*} 
 Similarly, for $\beta_k$ we have
\begin{equation*}
  \frac{\partial (-LL_k)}{\partial \beta_{k}} = -\frac{1}{\sigma_{k}} \bvec{r}_k^T V_{k}^T(\bvec{x}).
\end{equation*}
The derivative with respect to each $\gamma_j$, for $j \in \parents{k}$,  must be computed recursively. Consider
\begin{equation*}
  \frac{\partial (-LL_k)}{\partial \gamma_{j}} =  -\frac{1}{\sigma_k} \bvec{r}_k^T \diag\left(W_{jk}^T(\bvec{x})\alpha_{jk}\right) \frac{\partial f_{j}(\bvec{x}; \gamma_j)}{\partial \gamma_{j}} =  \bvec{p}^T_{jk} \frac{\partial f_{j}(\bvec{x}; \gamma_j)}{\partial \gamma_{j}},
\end{equation*}
where we have abused notation by letting $\frac{\partial f_{j}(\bvec{x}; \gamma_j)}{\partial \gamma_{j}}$ refer to the derivative of $f_j$ with respect to all parameters in $\gamma_j$ and 
\begin{equation*}
  \bvec{p}^T_{jk} = -\frac{1}{\sigma_k} \bvec{r}_k^T \diag\left(W_{jk}^T(\bvec{x})\alpha_{jk}\right)
\end{equation*}
denotes the chain rule information that needs to be propagated ``backward'' from node $k$ to node $j$. Since
\begin{equation*}
  f_{j}(\bvec{x};\gamma_{j}) = \sum_{\ell \in \parents{j}} f_{\ell}(\bvec{x}; \gamma_{\ell}) W_{\ell j}^T(\bvec{x}) \alpha_{\ell j} + V_{j}^T(\bvec{x})\beta_{j},
\end{equation*}
where $\gamma_\ell \subset \gamma_j$, we obtain the following expressions
\begin{align}
  \frac{\partial f_{j}(\bvec{x})}{\partial \alpha_{\ell j}} = f_{\ell}(\bvec{x}; \gamma_{\ell}) W_{\ell j}^T(\bvec{x}), \qquad
  \frac{\partial f_{j}(\bvec{x})}{\partial \beta_{j}} = V_{j}^T(\bvec{x}),  \qquad
  \frac{\partial f_{j}(\bvec{x})}{\partial \gamma_{\ell}} = \diag\left(W_{\ell j}^T(\bvec{x})\alpha_{\ell j} \right)\frac{\partial f_{\ell}(\bvec{x}; \gamma_\ell)}{\partial \gamma_{\ell}},
\end{align} 
where we see the third term provides the recursion. If $j$ were root \rev{node}, then only the middle term is needed. We can now repeat the process and compute all the gradients with respect to $f_{\ell}.$
Note that the gradient with respect to $\alpha_{\ell j}$ refers only to those $\alpha_{\ell j}$ in the parents of node $j$.

The overall pseudo-code for the forward sweep is provided by Algorithm~\ref{alg:forwardprop}, and the pseudo-code for the backward sweep is provided in Algorithm~\ref{alg:backprop}. In these algorithms, the symbol $(*)$ stands for element-wise multiplication and $(\otimes)$ is the Kronecker product.

\renewcommand{\algorithmicrequire}{\textbf{Inputs}}
\renewcommand{\algorithmicensure}{\textbf{Procedure}}
\begin{algorithm}[!t]
    \begin{algorithmic}[1]
      \REQUIRE \textrm{forward-sweep}(node $k$; inputs $\bvec{x}$; multifidelity surrogate $\mathcal{G}$)
      \STATE $\mathcal{A} = \ancestors{k} \cup \{k\}$
      \STATE $\mathcal{F} = \roots{\mathcal{G}} \cap \mathcal{A}$ ; relevant root nodes
      \STATE $queue = \textrm{FIFOQueue}()$ ;
      \FOR {$i \in \mathcal{A}$}
      \STATE $\partial z_{i} =  V_i^T(\bvec{x})$ ; partial gradient with respect to $\beta_i$
      \STATE $z_i = \partial z_{i} \beta_i$ ; evaluate
      \IF {$i \in \mathcal{F}$}
      \STATE $queue.put(i)$
      \ENDIF
      \ENDFOR
      \WHILE {$queue$ is not empty}
      \STATE $\ell = queue.get()$
      \FOR {$c \in \child{\ell}$ if $c \in \mathcal{A}$}
        \STATE $\partial z_{\ell c} =  \left(\bvec{1}_{1 \times p_{\ell c}} \otimes z_{\ell}\right) * W_{\ell c}^T(\bvec{x})$ ; partial gradient with respect to $\alpha_{\ell c}$
        \STATE $z_{\ell c} =  \partial z_{\ell c}  \alpha_{\ell c}$ 
        \STATE $z_{c} = z_{c} + z_{\ell c}$ \label{line:fw:up}
        \IF {$c$ has included all parents}
        \STATE $queue.put(c)$
        \ENDIF
      \ENDFOR
      \ENDWHILE
    \RETURN evaluations $z_{\ell}$ and partial gradients $\partial z_{\ell}$ and $\partial z_{ij}$ with respect to $\beta_{\ell}$ and $\alpha_{ij}$ for all $\ell,i,j \in \mathcal{A}$
    \end{algorithmic}
  \caption{Forward sweep and derivative precomputation}
  \label{alg:forwardprop}
\end{algorithm}

\subsection{Forward evaluation}\label{sec:forward_pass}
In this section we describe the forward sweep Algorithm~\ref{alg:forwardprop} and its computational cost. This algorithm evaluates all the ancestors of node $k$, at location $\mathbf{x}$. It also precomputes the quantities that will be used by the chain rule backward sweep to complete the derivative computation. For this discussion we assume that the size of all parameters $\alpha_{ij}$ and $\beta_i$ are at most $p$, there are $N$ data points, and the cost of a single basis computation $V_i^T(x)$ is some function $E(p)$ of the number of parameters.

The forward sweep begins by determining all the ancestors $\mathcal{A}$, all the nodes which are required to evaluate the final $f_k$. The determination of all ancestors for each node can be done prior to any training (it is part of the graph structure), and is considered an \textit{offline} cost. The algorithm then iterates through all of the ancestors and computes the basis matrix $V_i(\bvec{x})$ and resulting evaluation ---  a total cost of $\mathcal{O}(nE(p))$ for the basis function evaluation and $\mathcal{O}(np)$ for the matrix multiplication. The ancestor nodes that are also root nodes of the graph are then put into a first-in-first-out (FIFO) queue, which has $\mathcal{O}(1)$ access and retrieval.

The second part of the forward sweep is an iteration until the queue empties. Since we use a FIFO queue, this is a breadth-first algorithm. A node is removed from the queue, and all the children of that node are then considered. For each child, the derivative $\partial z_{\ell c}$ is computed, this quantity will be used to obtain the derivative with respect to $\alpha_{\ell c}$ in the backwards pass, \rev{and is} also used to update $z_{\ell c}$. Finally, on Line~\ref{line:fw:up} of Algorithm~\ref{alg:forwardprop} the evaluation $z_c$ is updated with the current parent. The asymptotic cost of each (and therefore all) of these three lines is $\mathcal{O}(nE(p)$),  Finally, if the child has considered all of its parents, it is entered into the queue. 
Suppose that the maximum number of children any node has is $C$ and that there are $A$ ancestors --- then the final asymptotic cost of the evaluation is $\mathcal{O}(nACE(p)).$ Here we see that the network structure critically affects the computational complexity of the evaluation. Sparser networks \revv{(i.e. less edges between nodes)} have less children for each node, and therefore incur smaller costs.

\subsection{Backward evaluation}\label{sec:backward_pass}
The backward evaluation pass in Algorithm~\ref{alg:backprop} applies the chain rule in a breadth-first search from the target node to all the roots in its ancestry. With $k$ being the target node, the algorithm begins by computing the gradient with respect to $\beta_k$, and then computes the chain rule factor $\bvec{p}_{k}$ to pass to its ancestors. Line~\ref{line:alg:pass_all} of Algorithm~\ref{alg:backprop} initializes (to zero) the factors that each node passes down to its ancestors. The target node is then put into another FIFO queue and a loop over the queue is performed until it is empty.

In each iteration of the loop, the parents of the node $\ell$ are considered. The following operations are then performed on each parent: Line~\ref{line:alg:prop_back}, the chain rule factor that passes to its ancestors is updated; Line~\ref{line:alg:deriv}, the derivative of $\alpha_{c\ell}$ is updated through chain rule; and Line~\ref{line:alg:deriv_beta}, the derivative with respect to $\beta_c$ is updated. The cost of each inner loop is $\mathcal{O}(np).$ Since it has to be performed for every parent in the hierarchy, the total cost will be $\mathcal{O}(nACp).$ Following these updates, a check is performed to determine if a parent has been updated by all of its children. Once it has been updated with all of its children, it has a complete $\bvec{p}_c^T$ to pass back to its own ancestors and is added to the queue.

\renewcommand{\algorithmicrequire}{\textbf{Inputs}}
\renewcommand{\algorithmicensure}{\textbf{Procedure}}
\begin{algorithm}[!t]
    \begin{algorithmic}[1]
      \REQUIRE \textrm{backward-sweep}(residual $\bvec{r}$, node $k$, ancestors $\mathcal{A}$, evals $z_l$; partial gradients $\partial z_{\ell}$ and $\partial z_{ij}$; multifidelity surrogate $\mathcal{G}$
      \STATE $\partial z_{k} = -\frac{1}{\sigma^2_k} \bvec{r}^T \partial z_k$ ; Gradient w.r.t $\beta_{k}$
      \STATE $\bvec{p}^T_k = -\frac{1}{\sigma^2_k} \bvec{r}_k^T$ ; multiplicative part to pass to ancestors (chain rule)
      \STATE $\bvec{p}^T_{\ell} = 0$ for all $\ell \in \mathcal{A}$ ; initialize chain rule passing to zero \label{line:alg:pass_all}
      \STATE $queue = \textrm{FIFOQueue}()$ ;
      \STATE $queue.put(k)$
      \WHILE {$queue$ is not empty}
      \STATE $\ell = queue.get()$
      \FOR {$c \in \parents{\ell}$}
      \STATE $\bvec{p}^T_{c} \leftarrow \bvec{p}^T_c + \bvec{p}^T_{\ell} * z_{c\ell}$ ; update chain rule \label{line:alg:prop_back}
      \STATE $\partial z_{c\ell} \leftarrow \bvec{p}_\ell * \partial z_{c \ell}$ ; final derivative w.r.t $\alpha_{c\ell}$ \label{line:alg:deriv}
      \STATE $\partial z_{c} =  \left(\bvec{p}_{\ell} * z_{c\ell}\right)^T \partial z_{c}$ ; update derivative w.r.t $\beta_{c}$ \label{line:alg:deriv_beta}
      \IF {$c$ has included all children in $\mathcal{A}$}
        \STATE $queue.put(c)$
      \ENDIF
      \ENDFOR
      \ENDWHILE
    \RETURN derivatives $\partial z_{\ell}$ and $\partial z_{ij}$ with respect to $\beta_{\ell}$ and $\alpha_{ij}$ for all $\ell,i,j \in \mathcal{A} \cup \{k\}$
    \end{algorithmic}
  \caption{Backward sweep for derivative computation}
  \label{alg:backprop}
\end{algorithm}

\section{Experiments}\label{sec:experiments}

In this section we consider four numerical experiments to demonstrate the benefits and flexibility of the proposed approach to multifidelity surrogate development.  In each case, we describe the models considered, the networks used, and the comparison between the proposed network and a hierarchical network.

The first example~\ref{sec:three_model} provides an intuitive representation of the multifidelity surrogate and \rev{motivates} the potential benefits of encoding relationships beyond hierarchical. The second example~\ref{sec:noisy} expands upon the first one by (1) considering a larger ensemble of 9 models, (2) considering a noisy measurement process, and (3) demonstrating an ability to inject problem knowledge into the representation of a multifidelity surrogate. The third example~\ref{sec:thermal_block}, is a representative problem of diffusion PDEs. Here we compare three model structures and sample over thousands of realizations of the data to show that a majority of the time, the hierarchical structure is not optimal. Finally, we consider a problem from direct field acoustic testing~\ref{sec:dfat}, where we demonstrate both the flexibility in what can be considered multifidelity information sources and the benefits of regularization.

Each example uses the same training Algorithms~\ref{alg:forwardprop} and~\ref{alg:backprop} within an approximate Newton BFGS optimization routine available as part of SciPy. The code is available from the author's github page\footnote{\url{https://www.github.com/goroda}}.

\subsection{Three model example}\label{sec:three_model}
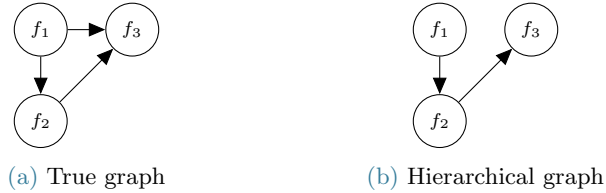
\begin{figure}
  \centering
  \tikzstyle{func} = [latent,  node distance=0.5cm, font=\scriptsize]
  \begin{subfigure}{0.3\textwidth}
    \centering
    \begin{tikzpicture}
      \node[func, ] at (0,0) (M1) {$f_{1}$} ;
      \node[func, below=of M1] (M2) {$f_{2}$};
      \node[func, right=of M1] (M3) {$f_3$};

      \edge {M1} {M2};
      \edge {M2} {M3};
      \edge {M1} {M3};
    \end{tikzpicture}
    \caption{True graph}
    \label{fig:three_model_true}
  \end{subfigure}
  ~
  \begin{subfigure}{0.3\textwidth}
    \centering
    \begin{tikzpicture}
      \node[func, ] at (0,0) (M1) {$f_{1}$} ;
      \node[func, below=of M1] (M2) {$f_{2}$};
      \node[func, right=of M1] (M3) {$f_3$};

      \edge {M1} {M2};
      \edge {M2} {M3};
    \end{tikzpicture}
    \caption{Hierarchical graph}
    \label{fig:three_model_rec}
  \end{subfigure}  
  \caption{True and hierarchical graphs considered for the model problem of Section~\ref{sec:three_model}, where we consider learning when the true graph structure is known.}
  \label{fig:three_model}
\end{figure}

In this section we demonstrate the benefit of accounting for non-hierarchical structure on a synthetic example with a known underlying graph. We consider the graphs shown in Figure~\ref{fig:three_model}. The graph in Figure~\ref{fig:three_model_true} is used to both generate the data and to fit the data. \rev{Physical models that can arise from these graphs are discussed at length in~\cite{Gorodetsky_JGE_2019}, and our aim here} is to show that when the underlying relationships amongst multiple models are known and not hierarchical, then we gain benefits from not using the predominant hierarchical approaches. The hierarchical graph for this case is shown in Figure~\ref{fig:three_model_rec}. This example seeks to illustrate that even though these two models can approximate functions of the same order (the longest chain has three nodes), there is an advantage to using a more relevant graph when data is limited.

\begin{figure}
  \centering
  \includegraphics[width=\textwidth]{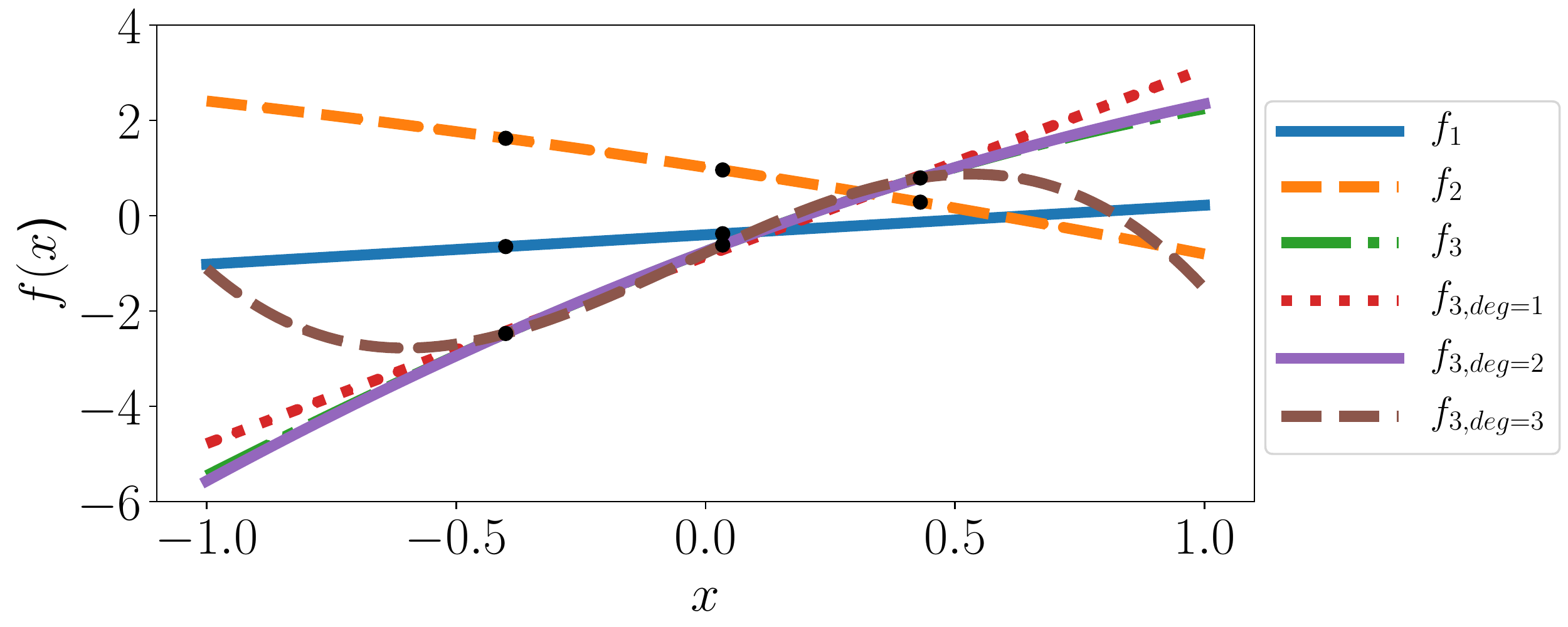}  
  \caption{Truth models, data, and single-fidelity regressions of the high-fidelity data for the model problem considered in Section~\ref{sec:three_model}.}
  \label{fig:true_funcs}
\end{figure}

The nodes and edges are parameterized by linear functions so that each of $\rho_{ij}$ and $\delta_{i}$ have two parameters (slope and y-intercept), 
\textit{i.e.} $\rho_{ij}(x) = \alpha_{ij,1} + \alpha_{ij,2} x$ and $\delta_i(x) = \beta_{i,1} + \beta_{i,2} x$. \revv{To generate the truth data, we randomly initialize the parameters of the non-hierarchical graph, these parameters are summarized in Table~\ref{tab:p1_true_params}.} This graph yields a high-fidelity model $f_3$ that is third order, however we restrict the high-fidelity data available during training to consist of only three data points for $f_3$. Two data points are used for $f_1$, and three data points are used for $f_2$. These data are nested, and they are shown along with their functions in Figure~\ref{fig:true_funcs}. In addition to these data points, we show reference regressions of first \revv{$f_{3,deg=1}$}, second \revv{$f_{3,deg=2}$}, and third \revv{$f_{3,deg=3}$} degree polynomials. None of these polynomials is able to recover $f_3$ since there is not enough data to adequately fit them in a single-fidelity context. 

\revv{Table~\ref{tab:p1_true_params} also shows the recovered parameters of the learned model. The parameters of $f_1(x) = \delta_1(x) = \beta_{1,1} + \beta_{1,2} x$ are recovered accurately, but the parameters of $f_2$ and $f_3$ less so. This is expected because the parameterization of the MFNets approximation is nonlinear and lacks uniqueness caused by the products between the functions $\rho_{ji}$ and $f_{i}$. As an example, consider the expanded equation for $f_2$
  \begin{align*}
    f_{2}(x) &= \rho_{12}(x)f_{1}(x) + \delta_{2}(x) \\
    &= \left(\alpha_{12,1} + \alpha_{12,2}x\right)\left(\beta_{1,1} + \beta_{1,2}x\right) + \beta_{2,1} + \beta_{2,2}x \\
    &= \left(\alpha_{12,1}\beta_{1,1} + \beta_{2,1}\right) +  \left(\alpha_{12,1} \beta_{1,2} + \alpha_{12,2}\beta_{1,1} + \beta_{2,2}\right)x + \alpha_{12,2}\beta_{1,2} x^2.
  \end{align*}
  From the last equality, which expresses  $f_2$ as a weighted sum of monomial of increasing degree, the lack of identifiability is evident. There can exist multiple combinations of $\alpha_{12}$ and $\beta_{i}$ that produce the same three scalar multipliers of the monomial terms. For example,  using the true network parameters the value of the constant monomial term coefficient is $\alpha_{12,1}\beta_{1,1} + \beta_{2,1} = 1.00379675$. A similar value $1.01479675$ is obtained using the learned network parameters. This value is very accurate despite the learned network parameters differing significantly. Similarly, the coefficients of the linear and quadratic monomial terms are recovered highly accurately. Thus we do recover the correct network, despite not recovering the exact parameters. In constrast, the root node is not over paramtererized, which suggests why it is recovered accurately.}

\begin{figure}
  \centering
  \begin{subfigure}{0.49\textwidth}
    \centering
    \includegraphics[width=\textwidth, clip, trim=0 0 100pt 0]{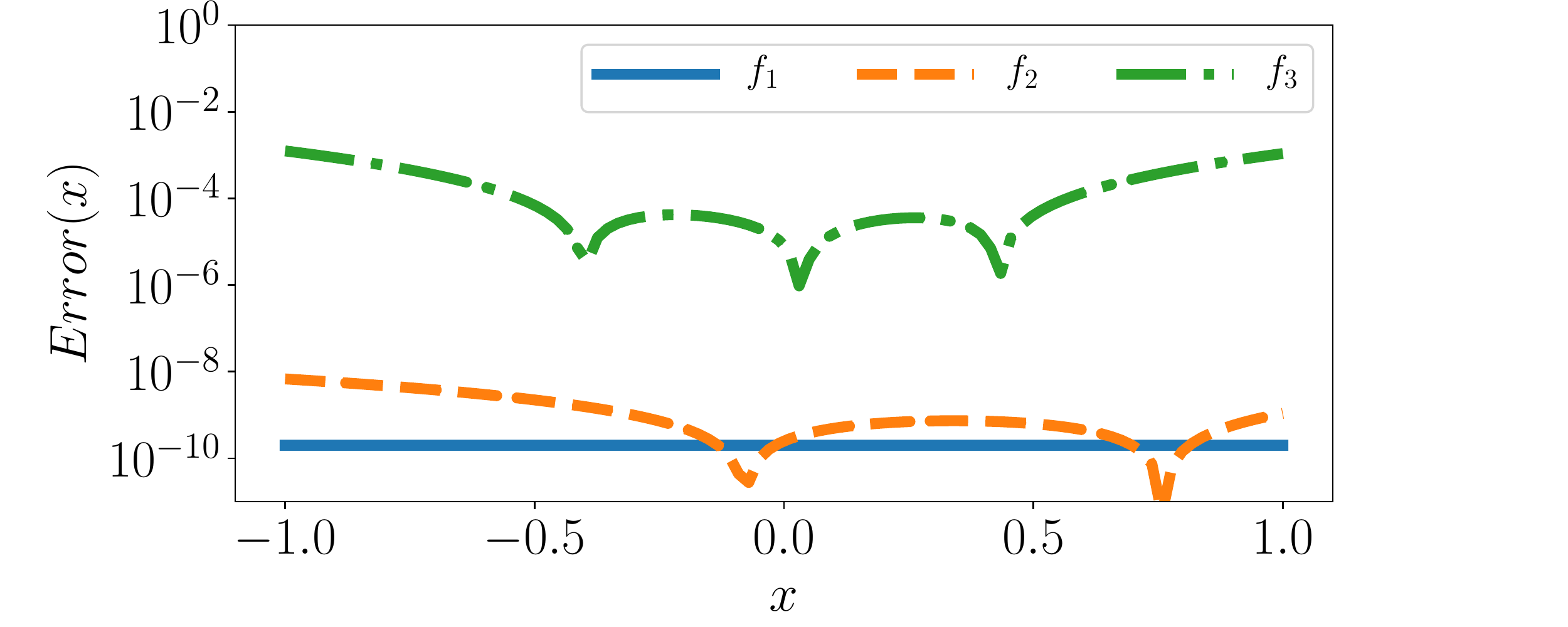}
    \caption{Recovery errors using the true network~\ref{fig:three_model_true}.}
  \end{subfigure}
  ~
  \begin{subfigure}{0.49\textwidth}
    \centering
    \includegraphics[width=\textwidth, clip, trim=0 0 100pt 0]{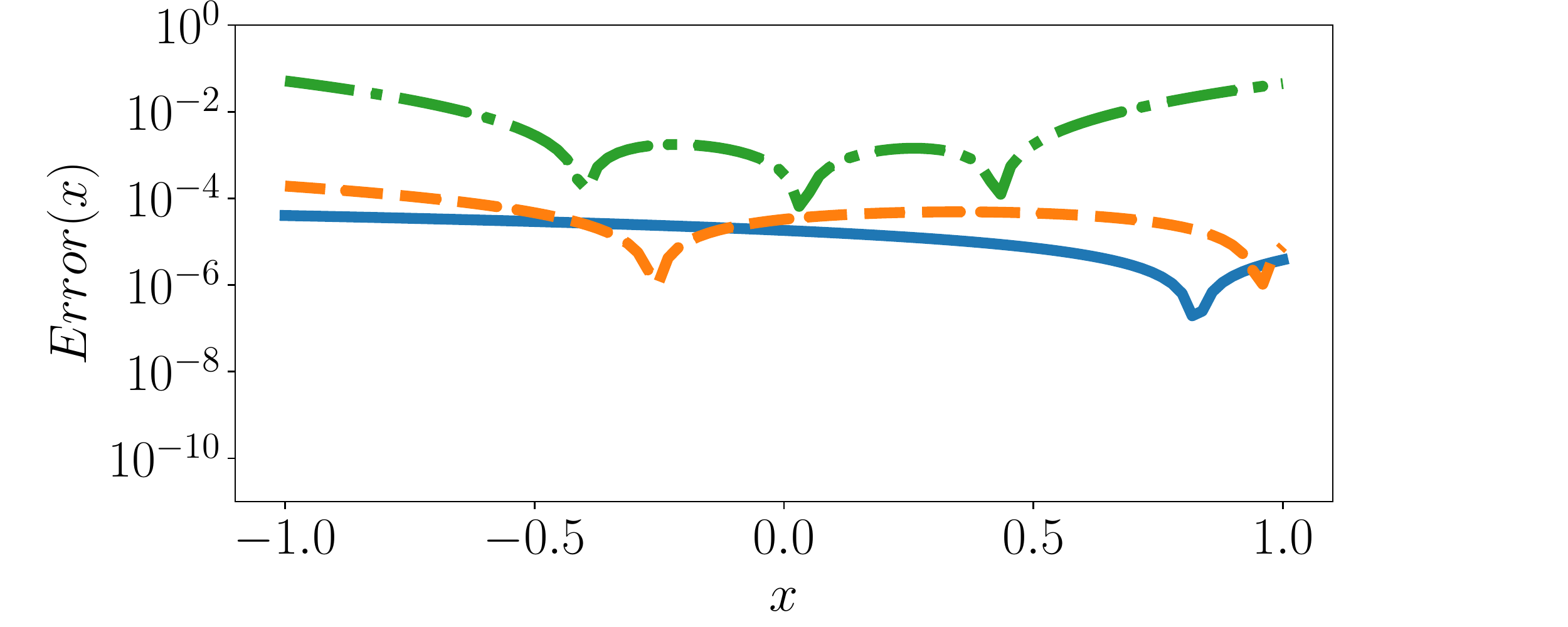}
    \caption{Recovery errors using the hierarchical network~\ref{fig:three_model_rec}.}
  \end{subfigure}
  \caption{Pointwise errors of \revv{the high-fidelity $f_1$ and low-fidelity $f_2$ and $f_3$ models obtained by regressing on the data using the true (data generating) and hierarchical networks. We observe an order of magnitude benefit in reconstructing $f_3$ (the high-fidelity model) using the true network compared with the hierarchical network --- indicating that leveraging the true structure can result in improved function recovery. Reconstruction quality of the two lower fidelity models is also vastly improved when using the true generative network.}}
  \label{fig:regress_three_model}
\end{figure}

 Figure~\ref{fig:regress_three_model} confirms that the pointwise reconstruction errors of these models are quite small. Figure~\ref{fig:regress_three_model} also shows the reconstruction error of the hierarchical model. We observe an order of magnitude benefit in reconstructing $f_3$ using the true network compared with the hierarchical network --- indicating that leveraging the true structure can result in improved function recovery.

\begin{table}
  \centering
  \caption{\revv{True and estimated parameters for the synthetic one dimensional model of Section~\ref{sec:three_model}. The estimated parameters are provided for the case when the true mode graph is being identified. The parameters $\beta_1$ for model $f_1$ are virtually identical, however the parameters involved in $f_2$ and $f_3$ differ. This does not mean that the predictive models learned are incorrect -- indeed from Figure~\ref{fig:regress_three_model} we see much smaller reconstruction errors. \revv{Rather the differences are due to the non-uniqueness the MFNETs parameterization}}}. \label{tab:p1_true_params}
  \begin{tabular}{|c|c|c|}
    \hline
    Parameter & True values [offset, slope] & Learned values [offset, slope]\\
    \hline    
    $\beta_1$ & $[-0.399999,   0.61917357]$ & $[-0.399999,    0.61917357]$\\
    $\beta_{2}$ & $[ 0.69834347, -1.25328053]$ & $[0.62987041, -1.1472885]$\\ 
    $\beta_{3}$ & $[0.45912744, 1.31524971]$ & $[0.62853275,  1.09869172]$\\
    $\alpha_{12}$ & $[-0.79113519, -0.34445981]$ & $[-0.96231826, -0.34445981]$\\
    $\alpha_{13}$ & $[-0.67351648, -0.32938732]$ & $[ 0.42886841, -0.25443088]$ \\
    $\alpha_{23}$ & $[-1.45728517,  0.59830806]$ & $[-1.18968888,  0.59172251]$\\
    \hline
  \end{tabular}
\end{table}

\subsection{Analytical noise example}\label{sec:noisy}
Next, we consider an analytical model which is not derived from a known graph. In this case, we also assume that the model is corrupted by Gaussian noise. Although synthetic, this example is inspired by applications where a model has both a functional dependence on uncertain parameters $x$, and a time-varying quantity whose statistics are difficult to estimate due to, for instance, a limited time-horizon. \revv{In these cases, evaluations of the quantity of interest are effectively only samples of the statistics of a long-running process, and can be therefore be treated as noisy samples from a numerical model.}
This scenario can occur when estimating averages of time-varying quantities in problems with unsteady dynamics and integration cannot occur indefinitely. \rev{For instance, it commonly arises in cases} of unsteady reacting turbulent flows where fluctuating quantities, \rev{\textit{e.g.}} temperature, pressure, velocities etc., are averaged to obtain their mean value in time possibly in correspondence of a point or a spatial domain. 

\revv{
Here, we choose a bivariate input $x \in \mathbb{R}^2$ 
and construct multifidelity surrogates using an ensemble of nine models
\begin{equation}
 \begin{split}
  f_k(x) = \left( 2 + (2 x_1^5 + 2 x_2^5) \Delta_1 + 3 x_1 x_2 + ( x_1^2 + x_2^2 + 5 x_1^2 x_2^2 ) \Delta_2 + 0.5 x_1 + 0.5 x_2 \right) \left( 1 + \mathbb{E}[ \mathcal{N}(0,1) ] \right), %
 \end{split}
 \label{eq:analytical}
\end{equation}
for $k=1,\dots,9$. This model ensemble consists of three model forms determined by the activation of model components by the values $\Delta_1, \Delta_2$ and additional three model forms due to three noise estimation levels $N$.} Specifically $\mathbb{E}[ \mathcal{N}(0,1)]$ is evaluated using the sample mean of $N$ observations, \textit{i.e.} $\frac{1}{N} \sum_{i=1}^N \mathcal{N}^{(i)}(0,1)$. The parameters for the nine models are reported in Table~\ref{tab: analytical_noise}.

\begin{table}[h!]
\centering
\begin{tabular}{ |c|c|c|c| } 
 \hline
       & $\Delta_1$ & $\Delta_2$ & N \\   
 \hline 
 $f_1$ & 0          & 0          & 5  \\  
 $f_2$ & 0          & 0          & 10  \\ 
 $f_3$ & 0          & 0          & 100  \\ 
 \hline
 $f_4$ & 0          & 1          & 5   \\ 
 $f_5$ & 0          & 1          & 10  \\ 
 $f_6$ & 0          & 1          & 100 \\ 
 \hline
 $f_7$ & 1          & 1          & 5   \\ 
 $f_8$ & 1          & 1          & 10  \\ 
 $f_9$ & 1          & 1          & 100 \\ 
 \hline
\end{tabular}
\caption{Analytical noise test case: parameters for the nine models. The highest fidelity model is $f_9$ as it includes all model terms and uses the most samples for estimating the expected value. The lowest fidelity model is $f_1$ as it includes the fewest model terms and the smallest number of samples for estimating the expectation.}
\label{tab: analytical_noise}
\end{table}

\begin{figure}
  \centering
  \begin{subfigure}{\textwidth}
    \centering
    \includegraphics[width=\textwidth]{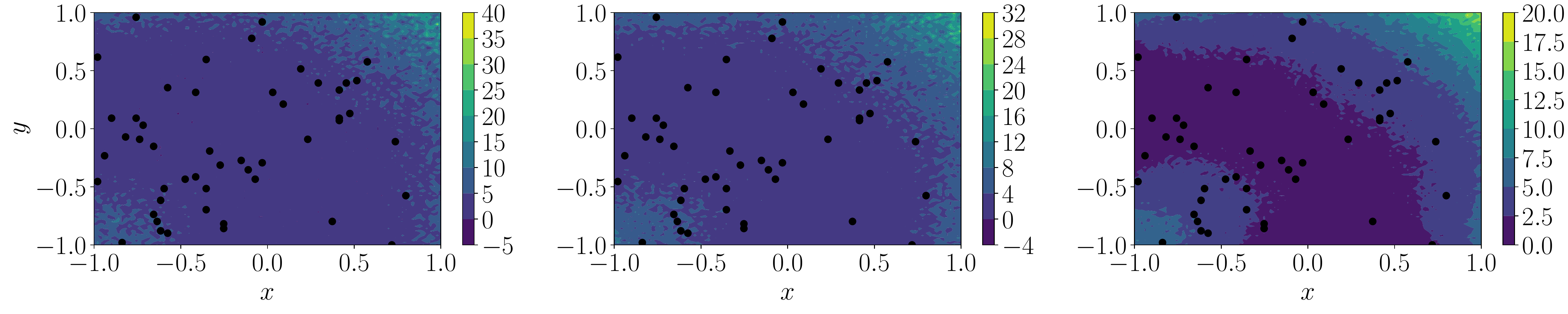}
    \caption{Left to right $f_{7}$, $f_{8}$, $f_{9}$}
  \end{subfigure}
  ~
  \begin{subfigure}{\textwidth}
    \centering
    \includegraphics[width=\textwidth]{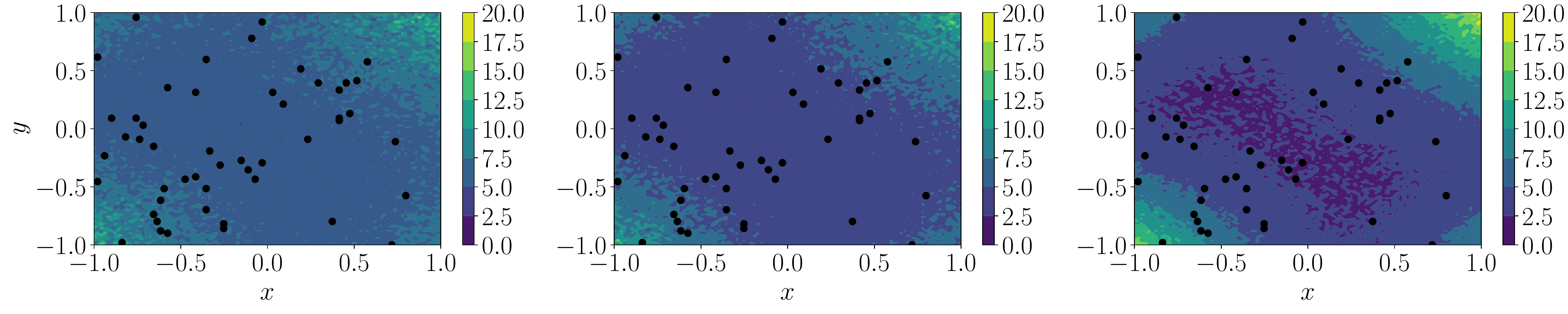}
    \caption{Left to right $f_{4}$, $f_{5}$, $f_{6}$}
  \end{subfigure}
  ~
  \begin{subfigure}{\textwidth}
    \centering
    \includegraphics[width=\textwidth]{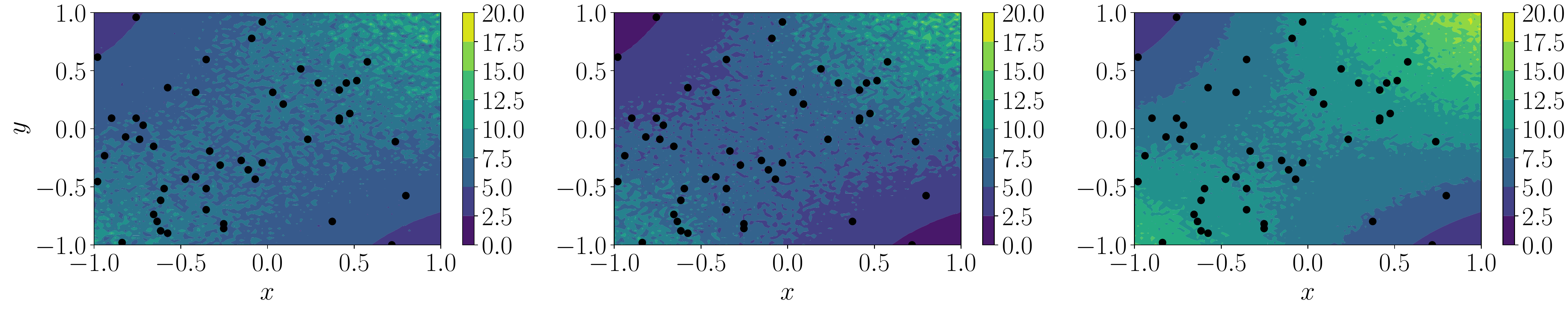}
    \caption{Left to right $f_{1}$, $f_{2}$, $f_{3}$}
  \end{subfigure}
  \caption{Analytical noise test case: responses for the nine models with locations of data}
  \label{fig:analytical_noise_responses}
\end{figure}

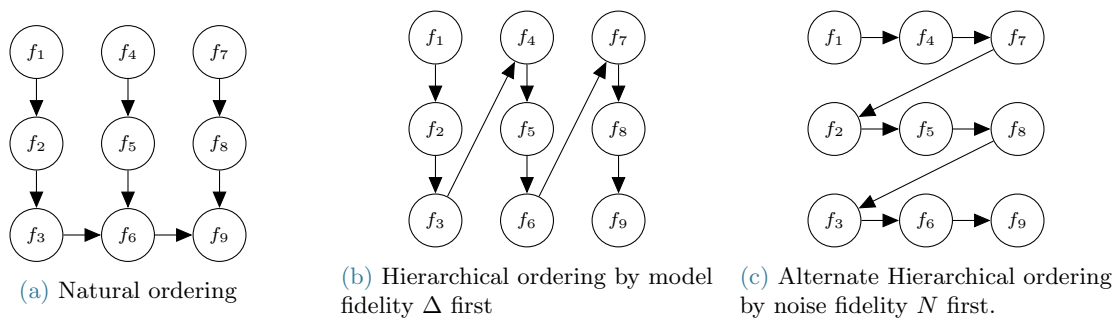
\begin{figure}
  \centering
  \tikzstyle{func} = [latent,  node distance=0.5cm, font=\scriptsize]
  \begin{subfigure}{0.3\textwidth}
    \centering
    \begin{tikzpicture}
      \node[func, ] at (0,0) (M1) {$f_{1}$} ;
      \node[func, below=of M1] (M2) {$f_{2}$};
      \node[func, below=of M2] (M3) {$f_3$};

      \node[func,  right=of M1] (M4) {$f_4$} ;
      \node[func, below=of M4] (M5) {$f_5$};
      \node[func, below=of M5] (M6) {$f_6$};

      \node[func, right=of M4] (M7) {$f_7$};
      \node[func, right=of M5] (M8) {$f_{8}$};
      \node[func, right=of M6] (M9) {$f_{9}$};

      \edge {M1} {M2};
      \edge {M2} {M3};
      \edge {M3} {M6};

      \edge {M4} {M5};
      \edge {M5} {M6};
      \edge {M6} {M9}; 
      
      \edge {M7} {M8};
      \edge {M8} {M9};
    \end{tikzpicture}
    \caption{Natural ordering}
  \end{subfigure}
  ~
  \begin{subfigure}{0.3\textwidth}
    \centering
    \begin{tikzpicture}
      \node[func, ] at (0,0) (M1) {$f_{1}$} ;
      \node[func, below=of M1] (M2) {$f_{2}$};
      \node[func, below=of M2] (M3) {$f_3$};

      \node[func, right=of M1] (M4) {$f_4$} ;
      \node[func, below=of M4] (M5) {$f_5$};
      \node[func, below=of M5] (M6) {$f_6$};

      \node[func, right=of M4] (M7) {$f_7$};
      \node[func, right=of M5] (M8) {$f_{8}$};
      \node[func, right=of M6] (M9) {$f_{9}$};

      \edge {M1} {M2};
      \edge {M2} {M3};
      \edge {M3} {M4};
      \edge {M4} {M5};
      \edge {M5} {M6};
      \edge {M6} {M7};
      \edge {M7} {M8};
      \edge {M8} {M9}; 
    \end{tikzpicture}
    \caption{Hierarchical ordering by model fidelity $\Delta$ first}
  \end{subfigure}
  ~
  \begin{subfigure}{0.3\textwidth}
    \centering
    \begin{tikzpicture}
      \node[func, ] at (0,0) (M1) {$f_{1}$} ;
      \node[func, below=of M1] (M2) {$f_{2}$};
      \node[func, below=of M2] (M3) {$f_3$};

      \node[func, right=of M1] (M4) {$f_4$} ;
      \node[func, below=of M4] (M5) {$f_5$};
      \node[func, below=of M5] (M6) {$f_6$};

      \node[func, right=of M4] (M7) {$f_7$};
      \node[func, right=of M5] (M8) {$f_{8}$};
      \node[func, right=of M6] (M9) {$f_{9}$};

      \edge {M1} {M4};
      \edge {M4} {M7};
      \edge {M7} {M2};
      \edge {M2} {M5};
      \edge {M5} {M8};
      \edge {M8} {M3};
      \edge {M3} {M6};
      \edge {M6} {M9};
    \end{tikzpicture}
    \caption{Alternate Hierarchical ordering by noise fidelity $N$ first.}
  \end{subfigure} 
  \caption{Analytical noise test case: models' natural structure versus \revv{two candidate hierarchical orderings.}}
  \label{fig:analytical_noise_structure}
\end{figure}

\begin{figure}
  \centering
  \includegraphics[width=\textwidth, clip, trim=0 200 0 10]{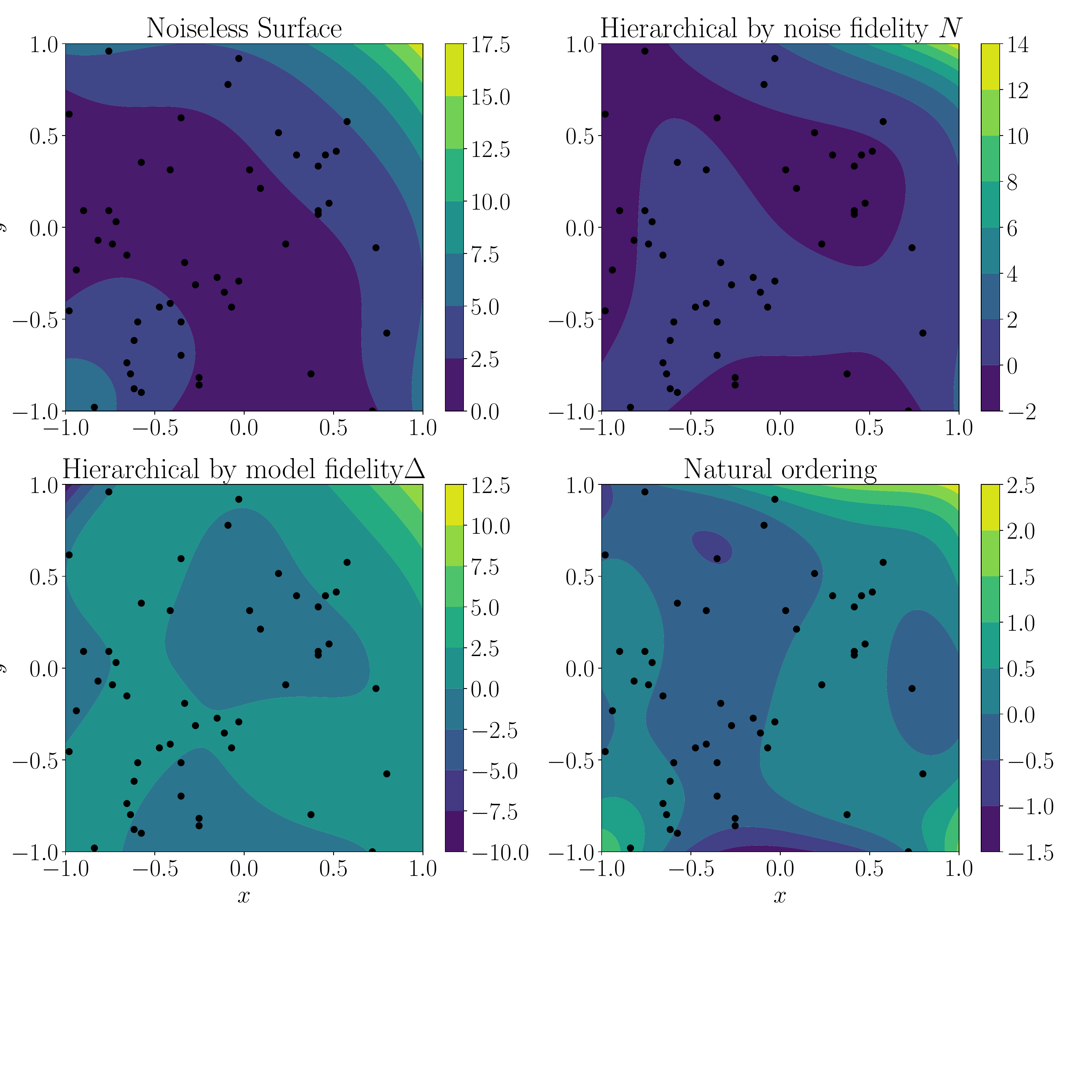}  
  \caption{\revv{Analytical noise test case: true noiseless surface (upper left); pointwise reconstruction errors for the hierarchical by $\Delta$ (lower left) and by noise (upper right) graphs; and pointwise errors for the natural graph (lower right). Note that the colorbars are different in each plot to display the magnitudes of the signal clearly. The natural graph has pointwise errors an order of magnitude lower than either of the hierarchical alternatives.}}
  \label{fig:analytical_noise_reconstruction}
\end{figure}

The response surfaces of the nine models, \revv{and the samples used to train surrogates}, are depicted in Figure~\ref{fig:analytical_noise_responses}.
\revv{In this example, unlike the previous, we cannot determine the best multifidelity network representation of the nine models, 
t}herefore, we consider three different options: \revv{the ``natural'' ordering where the models are ordered according to the number of observations $N$; a hierarchical ordering where the nine models are ordered by their model-form fidelity $\Delta$ first and number of observations (noise) $N$ second; and an alternative hierarchical scheme in which they are ordered first by noise and then by $\Delta$. These three structures} are shown in Figure~\ref{fig:analytical_noise_structure}. The non-noisy high-fidelity model $f_9$ and the pointwise reconstruction error obtained by these two networks is shown in Figure~\ref{fig:analytical_noise_reconstruction}. Here we see that the natural (non-hierarchical) ordering is able to obtain an order of magnitude smaller errors than the hierarchical orderings.

\subsection{Thermal Block}\label{sec:thermal_block}
\begin{figure}[htb]
  \centering
  \includegraphics[width=0.3\textwidth]{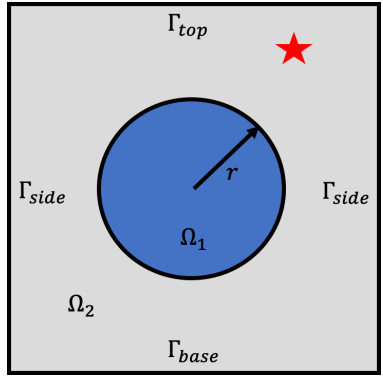}  
  \caption{Heat conduction in a two-dimensional domain $\Omega=\Omega_1\cup\Omega_2$. Measurements of the solution are taken at the point (0.5,0.8) depicted by the red star.}
  \label{fig:thermal_block}
\end{figure}

In this section we use multifidelity information fusion to predict steady-state heat diffusion in a two-dimensional domain $\Omega$ shown in Figure \ref{fig:thermal_block}. In this example our aim is to predict the temperature at a pointwise location $(0.5,0.8)\in\Omega=[-1,1]^2$, as a function of two parameters $\boldsymbol{\mu}=(\mu_0,\mu_1)$ which are independent uniform variables on $[0.1,10]$ and $[-1,1]$ respectively. The variable $\mu_0$ defines the diffusivity inside the circular inclusion and the variable $\mu_1$ controls the flux along the bottom boundary. This example was based upon a tutorial on constructing reduced order models using RBniCS \cite{HesthavenRozzaStamm2015}\footnote{\url{https://github.com/mathLab/RBniCS}}.

We consider three different approximations of the governing equations 
\begin{align}\label{eq:thermal}
\begin{cases}
    - \text{div} (\kappa(\mu_0)\nabla u(\boldsymbol{\mu})) = 0 & \text{in } \Omega,\\
    u(\boldsymbol{\mu}) = 0 & \text{on } \Gamma_{top},\\
    \kappa(\mu_0)\nabla u(\boldsymbol{\mu})\cdot \mathbf{n} = 0 & \text{on } \Gamma_{side},\\
    \kappa(\mu_0)\nabla u(\boldsymbol{\mu})\cdot \mathbf{n} = \mu_1 & \text{on } \Gamma_{base}.
\end{cases}
& &
\kappa(\mu_0) =
\begin{cases}
  \mu_0 & \text{in } \Omega_1,\\
  1 & \text{in } \Omega_2,\\
\end{cases}
\end{align}
These models include two finite element models with different mesh resolutions and a \rev{reduced-order} model. We construct all these models using RBniCS\cite{HesthavenRozzaStamm2015}. The \rev{highest-fidelity} information source $f_3$ uses linear finite elements with 1437 degrees of freedom. The first low-fidelity model $f_2$ uses linear finite elements with 186 degrees of freedom, and the last low-fidelity model $f_1$ is a two term reduced order model constructed using 20 snapshots of the high-fidelity information source. The normalized costs of evaluating each of these models for a single realization of the parameters $\boldsymbol{\mu}$ are 1,0.52, and 0.05 respectively.

\begin{figure}
  \centering
  \tikzstyle{func} = [latent, node distance=0.5cm, font=\scriptsize]
  \begin{subfigure}{0.3\textwidth}
    \centering
    \begin{tikzpicture}
      \node[func] at (0,0) (M3) {$f_3$};
      \node[func, above left=of M3] (M1) {$f_{1}$} ;
      \node[func, above right=of M3] (M2) {$f_{2}$};

      \edge {M1} {M2};
      \edge {M2} {M3};
      \edge {M1} {M3};
    \end{tikzpicture}
    \caption{Full}\label{fig:thermal-full-graph}
  \end{subfigure}
  ~
  \begin{subfigure}{0.3\textwidth}
    \centering
    \begin{tikzpicture}
      \node[func] at (0,0) (M3) {$f_3$};
      \node[func, above left=of M3] (M1) {$f_{1}$} ;
      \node[func, above right=of M3] (M2) {$f_{2}$};

      \edge {M2} {M3};
      \edge {M1} {M3};
    \end{tikzpicture}
    \caption{Peer}    
  \end{subfigure}
  ~
  \begin{subfigure}{0.3\textwidth}
    \centering
    \begin{tikzpicture}
      \node[func] at (0,0) (M3) {$f_3$};
      \node[func, above left=of M3] (M1) {$f_{1}$} ;
      \node[func, above right=of M3] (M2) {$f_{2}$};

      \edge {M2} {M3};
      \edge {M1} {M2};
    \end{tikzpicture}
    \caption{Hierarchical}\label{fig:thermal-hier-graph}
  \end{subfigure}    
  \caption{Thermal block model network structures.}
  \label{fig:thermal_block_structures}
\end{figure}
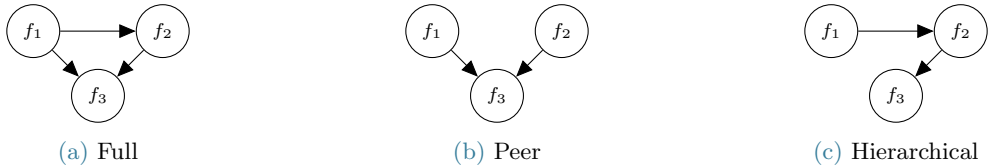

In the following we investigate the performance of our algorithm using the three graphs depicted in Figure~\ref{fig:thermal_block_structures}. Our aim is to show that for this problem, the non-hierarchical graphs typically perform better than the hierarchical graph. Each network model (depicted in Figure~\ref{fig:thermal_block_structures}) uses linear functions for the edges ($\rho_{ij}(x)$) and for nodes ($\delta_{i}$). As such the actual representative power of the hierarchical and full graphs is greater than the peer model --- they can represent third order polynomials, whereas the peer graph can only represent second order polynomials. Even so, the peer graph model outperforms the recursive model in this data regime.

\begin{figure}
  \centering
  \begin{subfigure}{0.32\textwidth}
    \includegraphics[width=\textwidth,clip,trim=18pt 10pt 20pt 10pt]{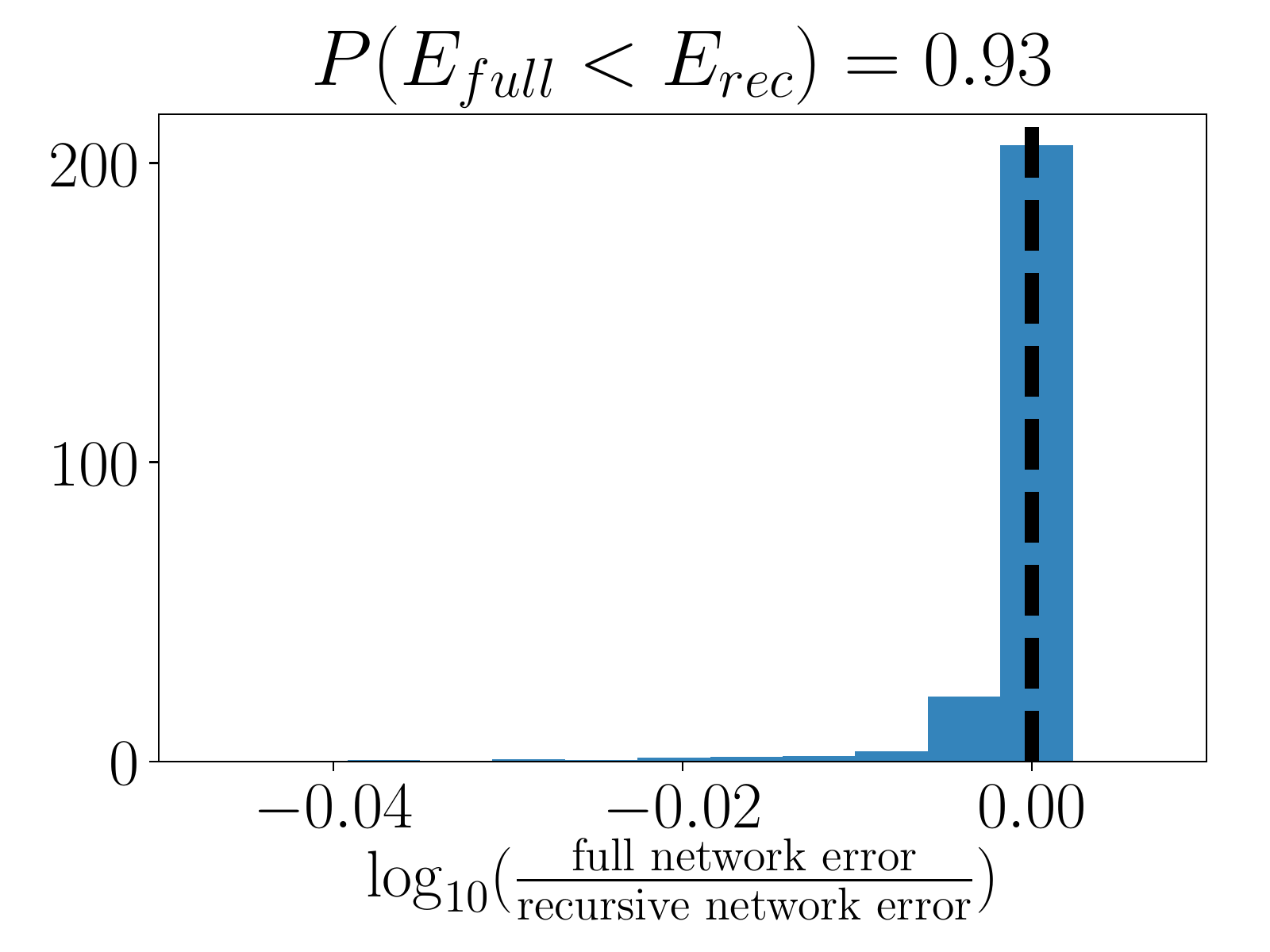}
    \caption{20/5/2}
  \end{subfigure}
  ~
  \begin{subfigure}{0.32\textwidth}
    \includegraphics[width=\textwidth,clip,trim=20pt 10pt 20pt 10pt]{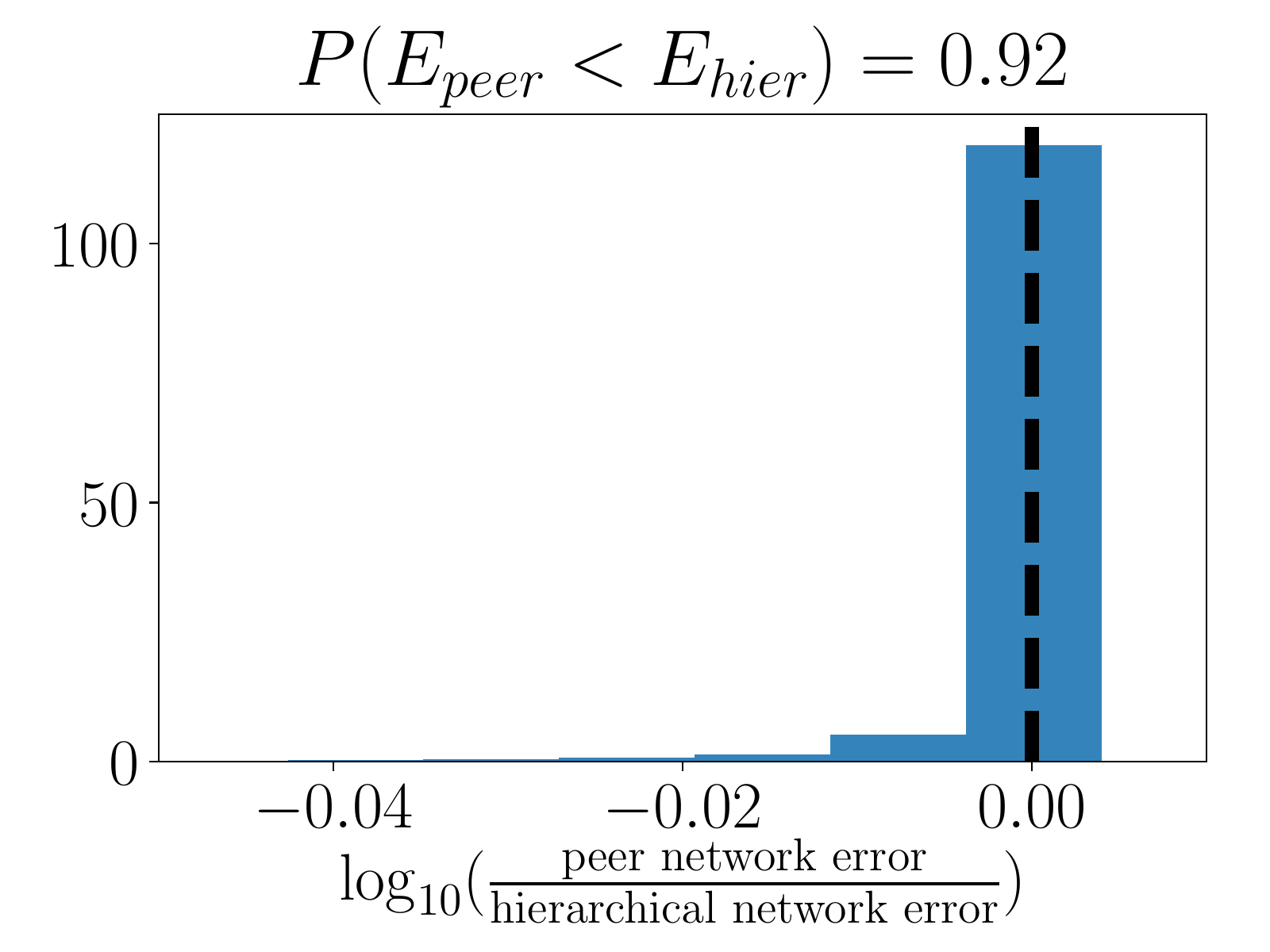}
    \caption{20/5/2}
  \end{subfigure}
  ~
  \begin{subfigure}{0.32\textwidth}
    \includegraphics[width=\textwidth,clip,trim=20pt 10pt 20pt 10pt]{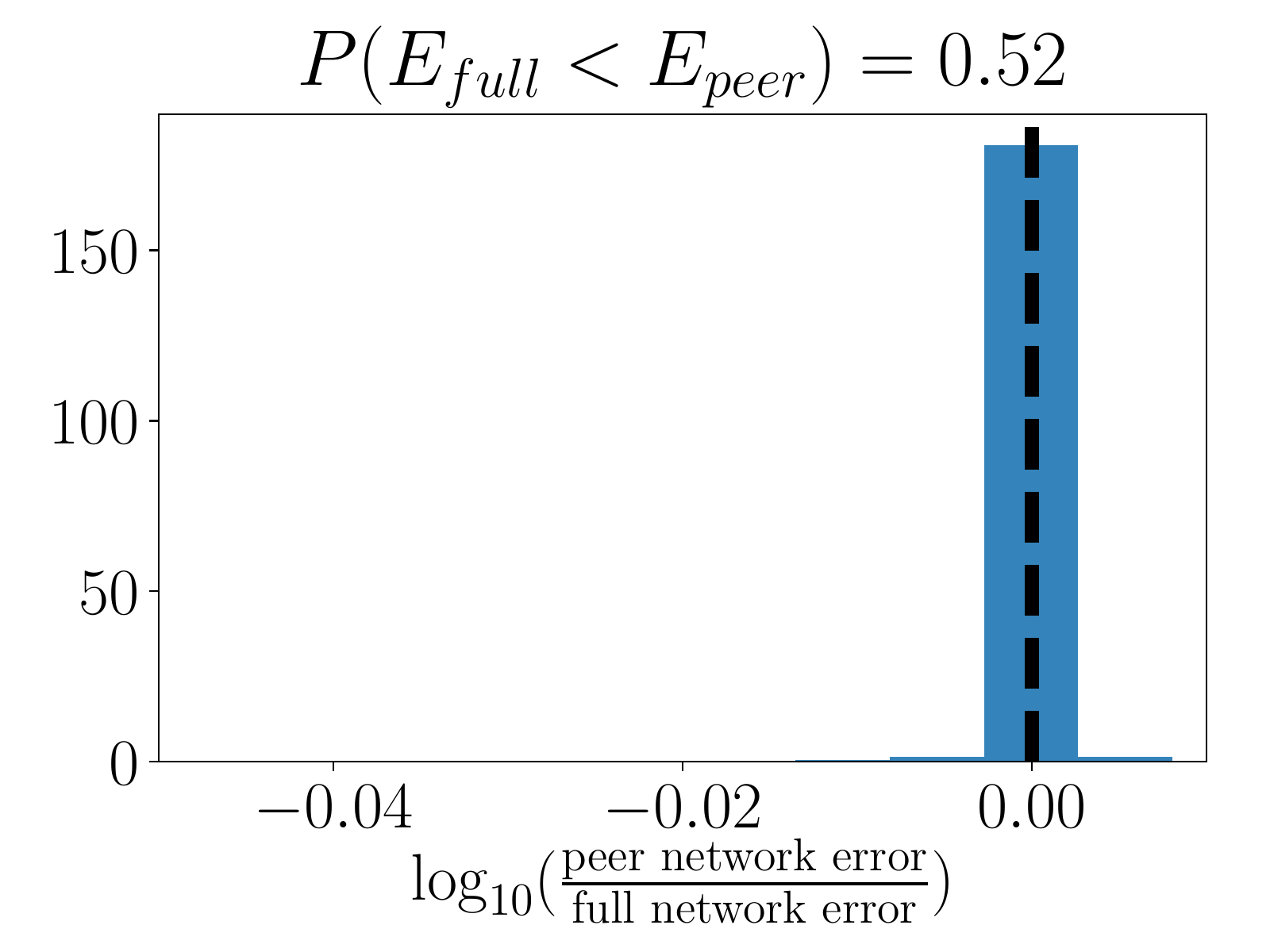}
    \caption{20/5/2}
  \end{subfigure}
  \begin{subfigure}{0.32\textwidth}
    \includegraphics[width=\textwidth,clip,trim=30pt 10pt 20pt 10pt]{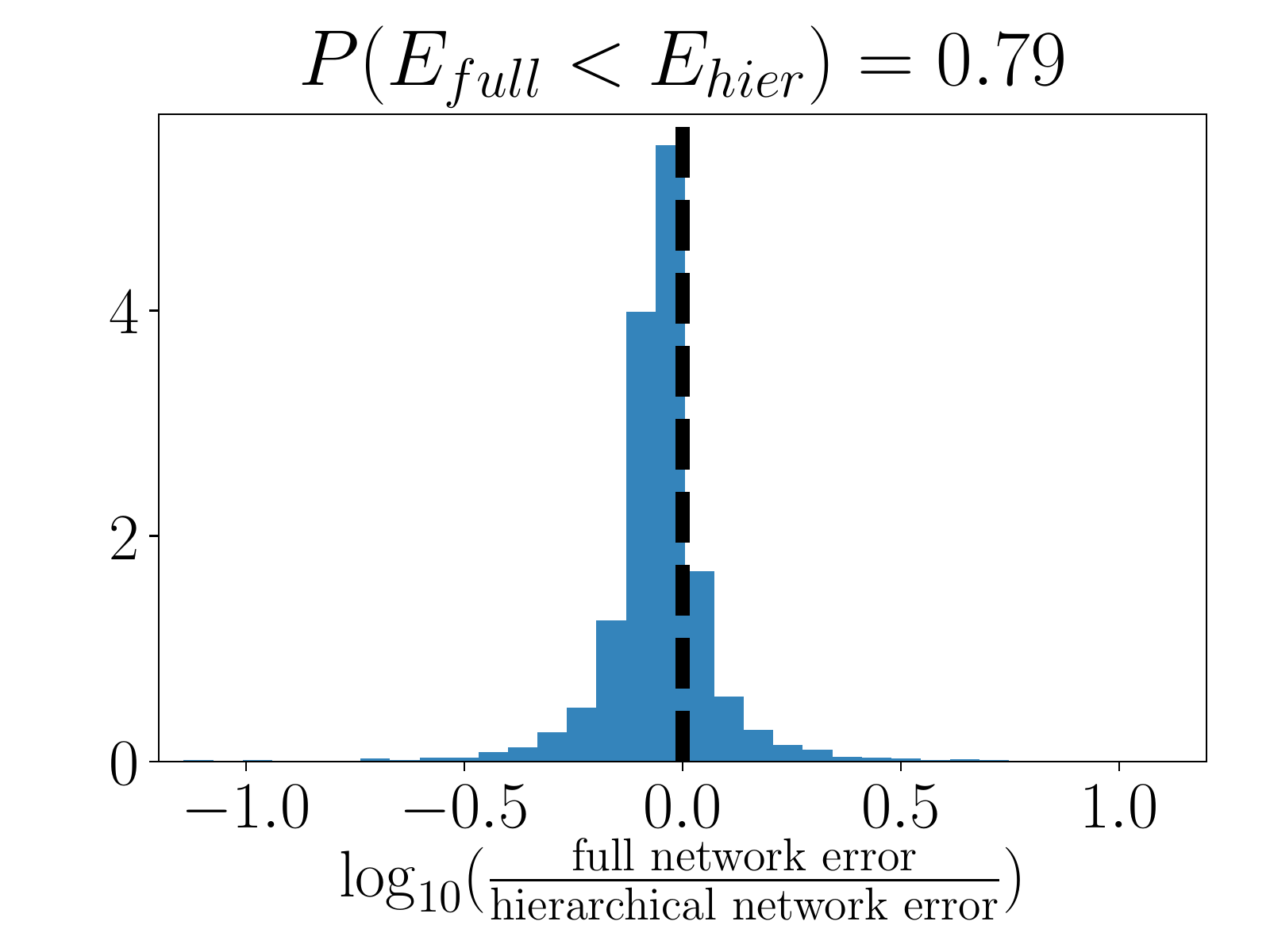}
    \caption{40/10/5}
  \end{subfigure}
  ~
  \begin{subfigure}{0.32\textwidth}
    \includegraphics[width=\textwidth,clip,trim=30pt 10pt 20pt 10pt]{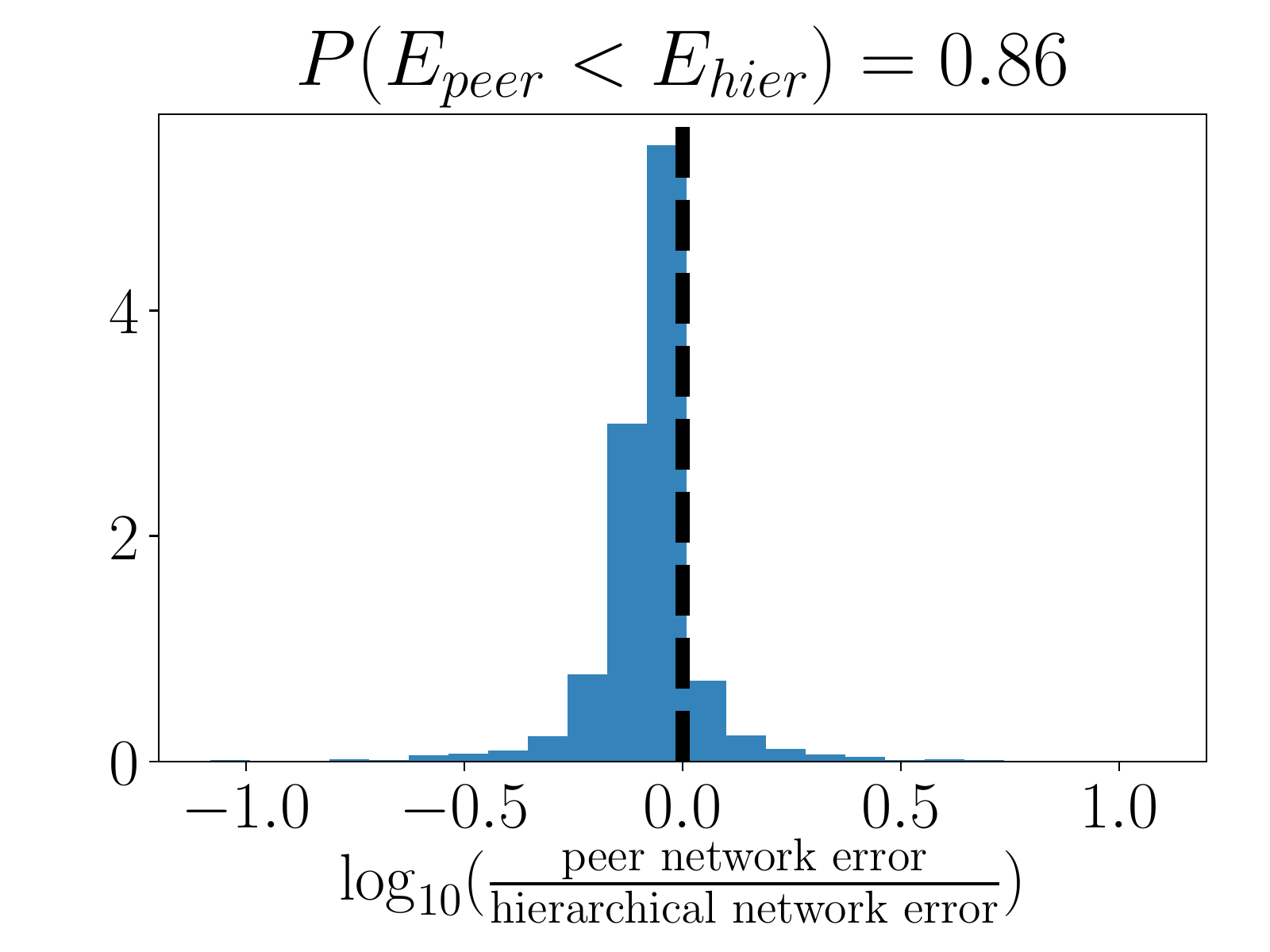}
    \caption{40/10/5}
  \end{subfigure}
  ~
  \begin{subfigure}{0.32\textwidth}
    \includegraphics[width=\textwidth,clip,trim=30pt 10pt 20pt 10pt]{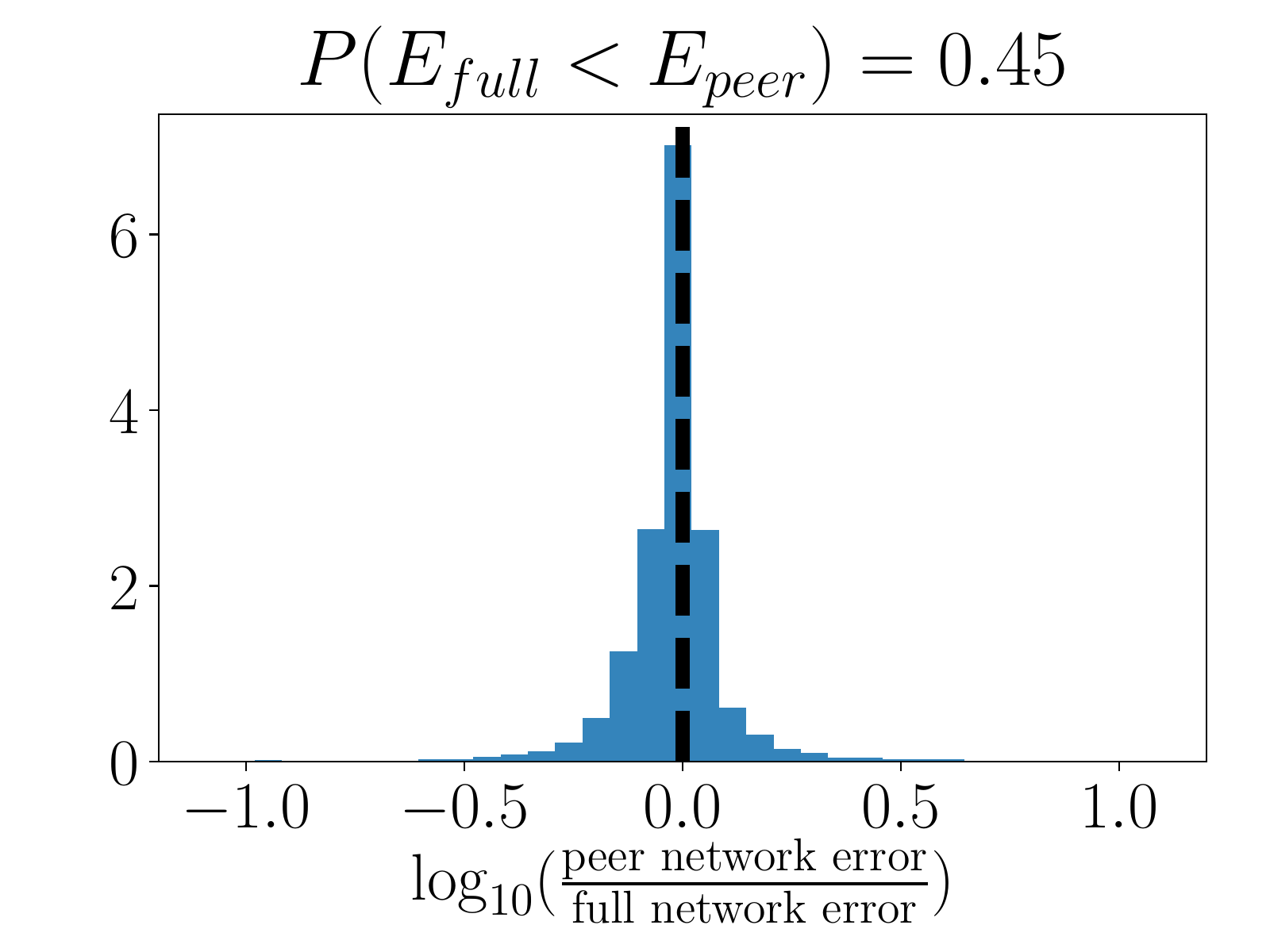}
    \caption{40/10/5}
  \end{subfigure}
  \begin{subfigure}{0.32\textwidth}
    \includegraphics[width=\textwidth,clip,trim=30pt 10pt 20pt 10pt]{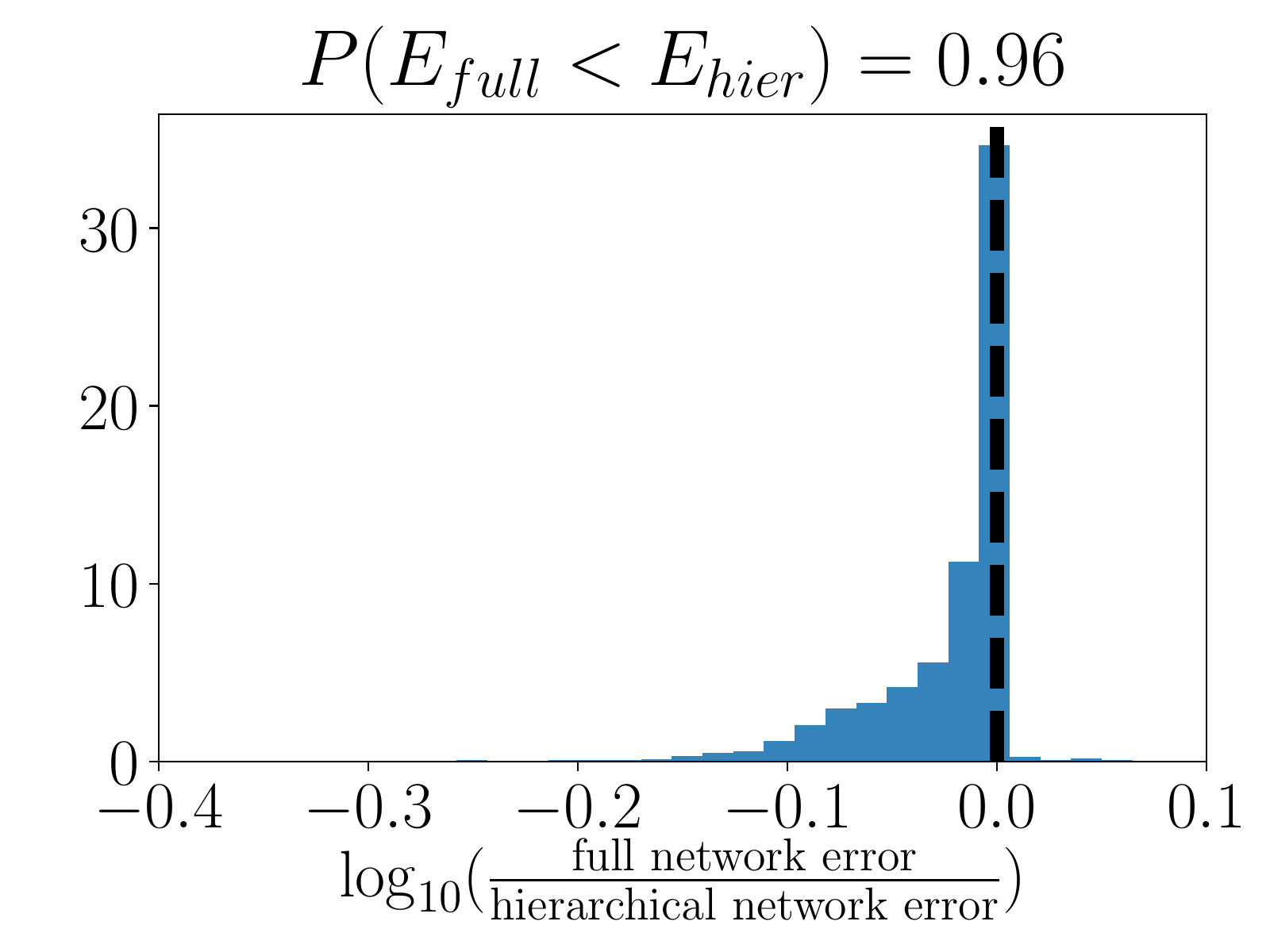}
    \caption{300/30/3}
  \end{subfigure}
  ~
  \begin{subfigure}{0.32\textwidth}
    \includegraphics[width=\textwidth,clip,trim=30pt 10pt 20pt 10pt]{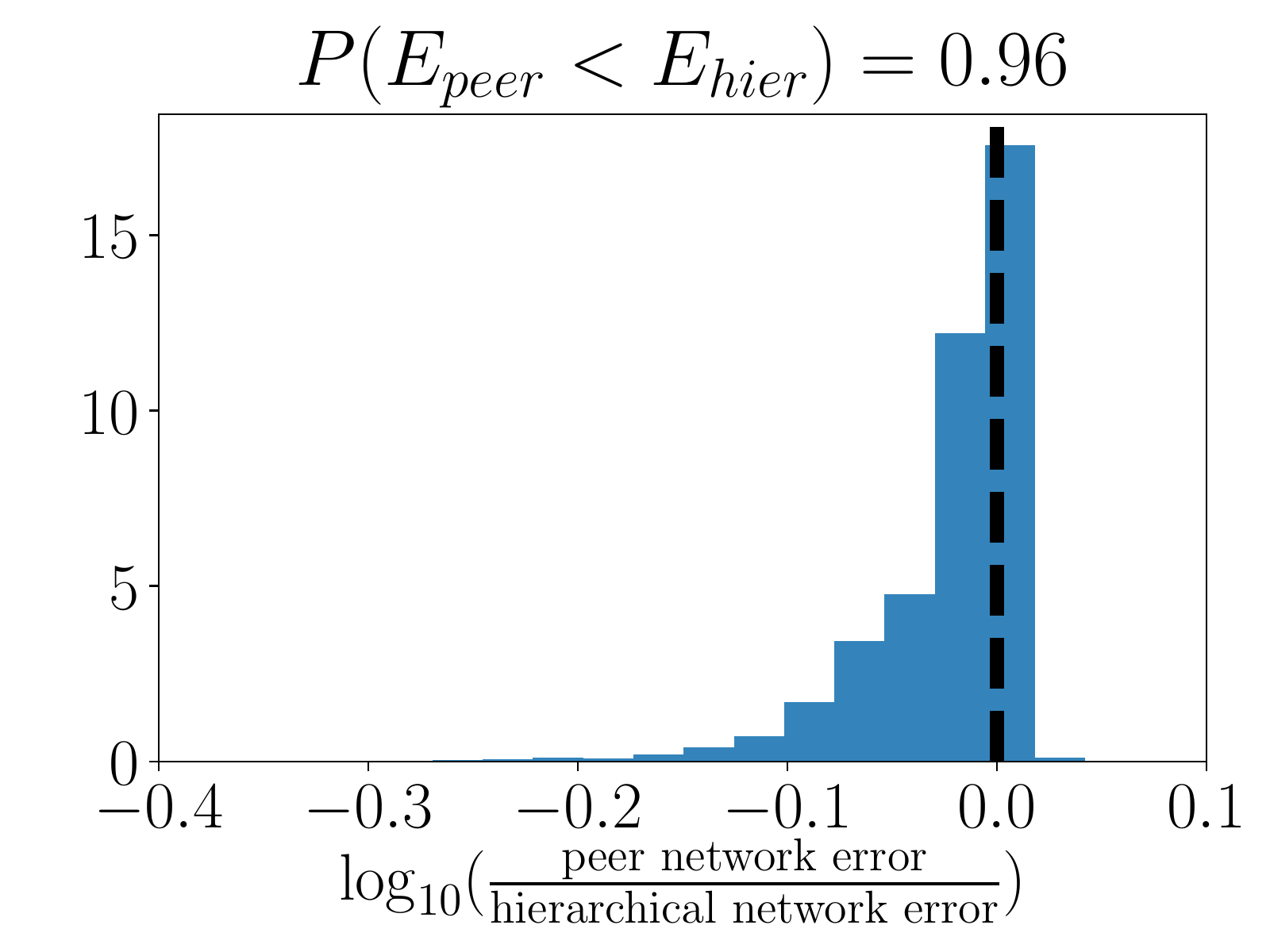}
    \caption{300/30/3}
  \end{subfigure}
  ~
  \begin{subfigure}{0.32\textwidth}
    \includegraphics[width=\textwidth,clip,trim=30pt 10pt 20pt 10pt]{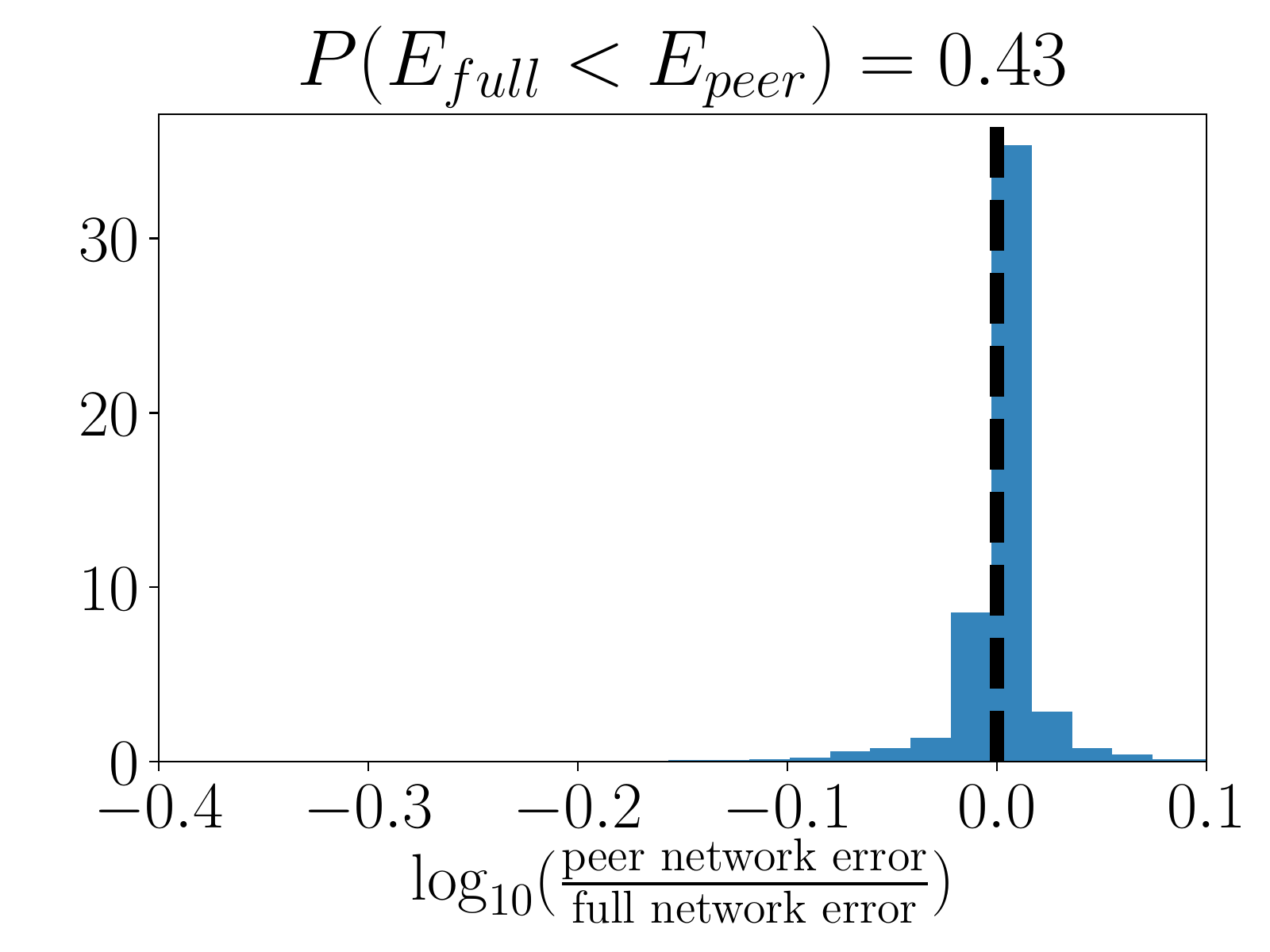}
    \caption{300/30/3}
  \end{subfigure}    
  \caption{Pairwise test-set error comparisons between different networks for different training data sizes on the thermal block problem. First column corresponds to comparison between full and hierarchical network, second column corresponds to comparison between peer and hierarchical, and third column corresponds to comparison between full and peer. First row corresponds to 20/5/2 training set distribution between low, medium, and high-fidelity models. Middle corresponds to 40/10/5, and last to 300/30/3. Each histogram is built using 5000 realization of training/testing data. The vertical dashed black line indicates boundary at which the networks perform equally. Both peer and full models consistently outperform the hierarchical model. Improved \revv{relative} performance is greater when there is \revv{a lower number} of high-fidelity data.}
  \label{fig:model_error_comparisons}
\end{figure}

In Figure~\ref{fig:model_error_comparisons} we \revv{compare the accuracy of the three different multifidelity graphs for varying amounts of training data. Each row in the plot compares the three different graphs for a 3-tuple (low/medium/high provided in the subcaptions) specifying the number of training samples allocated to the 3 model fidelities. Each histogram depicts the ratio of the mean squared errors computed using two different graphs (listed in the subplot title) and 5000 different draws of training data from a candidate set of 1000 samples; error is computed using the samples not used for training of withheld testing data.}
We find that the non-hierarchical graphs outperform the hierarchical graphs in the low-data (for the high-fidelity model) regime. These results empirically reinforce our hypothesis that exploiting the correct structure yields higher data-efficiency.

\subsection{Direct Field Acoustic Testing}
\label{sec:dfat}

In this section we use our multifidelity information fusion to fuse multiple direct field acoustic testing (DFAT) experiments that characterize performance of engineered structures under extreme vibration environments~\cite{ECStasinunas_RASchultz_MRRoss_2016a}. We also present the benefits of using regularization, and specifically sparse regularization, for \revv{identifying active and non-active edges in the graph.}

\begin{figure}[!ht]
  \begin{center}
  \begin{minipage}{0.24\textwidth}
    \centering
    {\bf Domain}\\[7pt]
    \begin{tikzpicture}[scale=1.1]
      \draw[very thick] (-0.6213,-1.5) -- ( 0.6213,-1.5);
      \draw[very thick] ( 0.6213,-1.5) -- ( 1.5,-0.6213);
      \draw[very thick] ( 1.5,-0.6213) -- ( 1.5, 0.6213);
      \draw[very thick] ( 1.5, 0.6213) -- ( 0.6213, 1.5);
      \draw[very thick] ( 0.6213, 1.5) -- (-0.6213, 1.5);
      \draw[very thick] (-0.6213, 1.5) -- (-1.5, 0.6213);
      \draw[very thick] (-1.5, 0.6213) -- (-1.5,-0.6213);
      \draw[very thick] (-1.5,-0.6213) -- (-0.6213,-1.5);

      \draw[very thick,green] (-0.5*0.6213,-1.5) -- ( 0.5*0.6213,-1.5);
      \draw[very thick,green] ( 1.5,-0.5*0.6213) -- ( 1.5, 0.5*0.6213);
      \draw[very thick,green] ( 0.5*0.6213, 1.5) -- (-0.5*0.6213, 1.5);
      \draw[very thick,green] (-1.5, 0.5*0.6213) -- (-1.5,-0.5*0.6213);

      \draw[very thick,green] ( 0.6213+0.25*0.8787,-1.5+0.25*0.8787) -- ( 0.6213+0.75*0.8787,-1.5+0.75*0.8787);
      \draw[very thick,green] ( 1.5-0.25*0.8787, 0.6213+0.25*0.8787) -- ( 1.5-0.75*0.8787, 0.6213+0.75*0.8787);
      \draw[very thick,green] (-0.6213-0.25*0.8787, 1.5-0.25*0.8787) -- (-0.6213-0.75*0.8787, 1.5-0.75*0.8787);
      \draw[very thick,green] (-1.5+0.25*0.8787,-0.6213-0.25*0.8787) -- (-1.5+0.75*0.8787,-0.6213-0.75*0.8787);

      \draw[very thick,fill=red]  ( 0,     0)     circle (0.5);
      \draw[very thick,fill=blue] (-0.177,-0.177) circle (0.25);
    \end{tikzpicture}
  \end{minipage}\hspace{1.25cm}
  \begin{minipage}{0.22\textwidth}
    \centering
    {\bf Pressure} \\[7pt]
    \makebox{\includegraphics[width=1.1\textwidth]{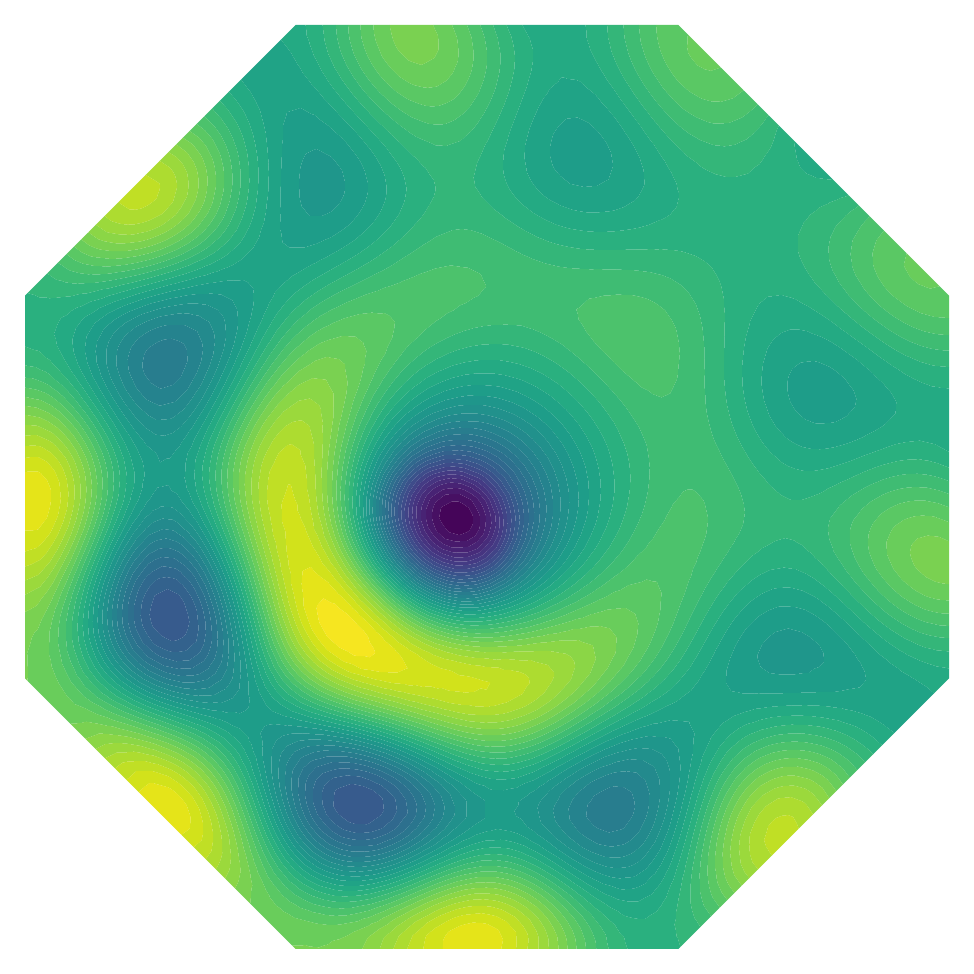}}
  \end{minipage}\hspace{1.25cm}
  \begin{minipage}{0.22\textwidth}
    \centering
    {\bf 1st Basis Function} \\[7pt]
    \makebox{\includegraphics[width=1.1\textwidth]{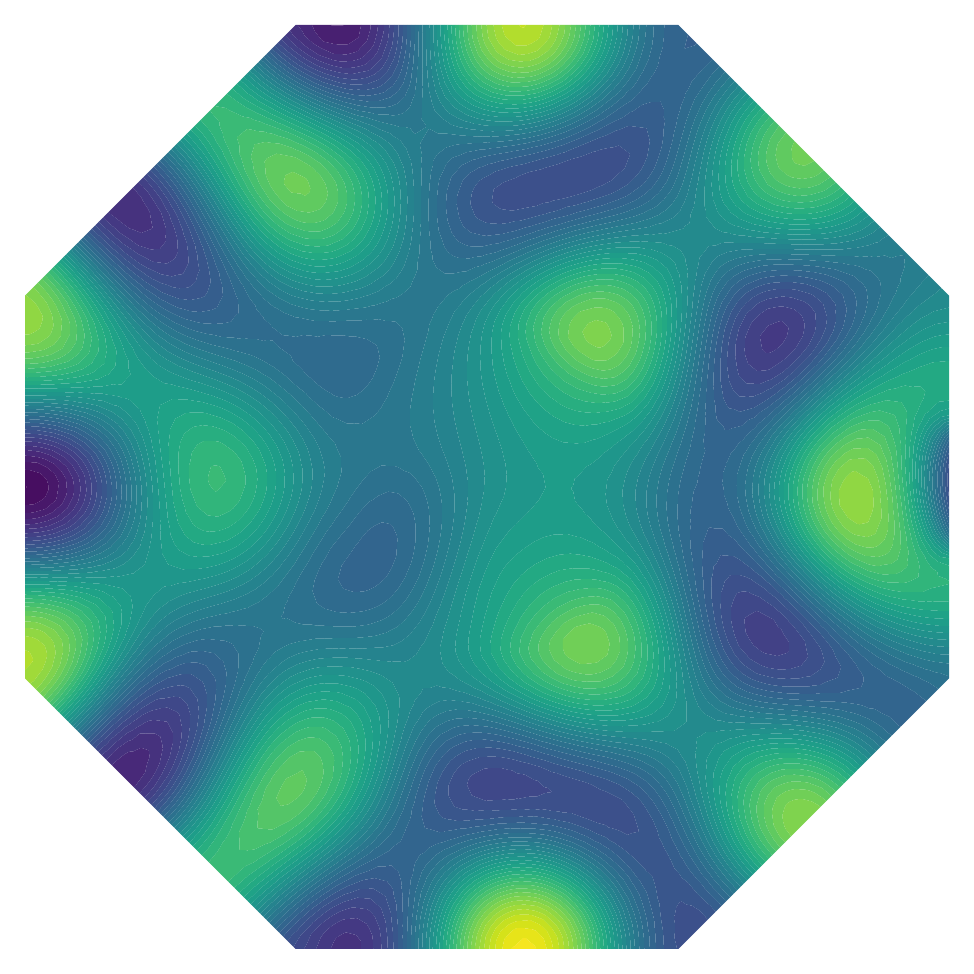}}
  \end{minipage}
  \end{center}
  \caption{Left to Right: The direct field acoustic testing experimental setup, the acoustic pressure of the high-fidelity information source, and the basis function obtained when activating the 1st speaker (directly to the right of the circular inclusion).}
  \label{fig:dfat}
\end{figure}
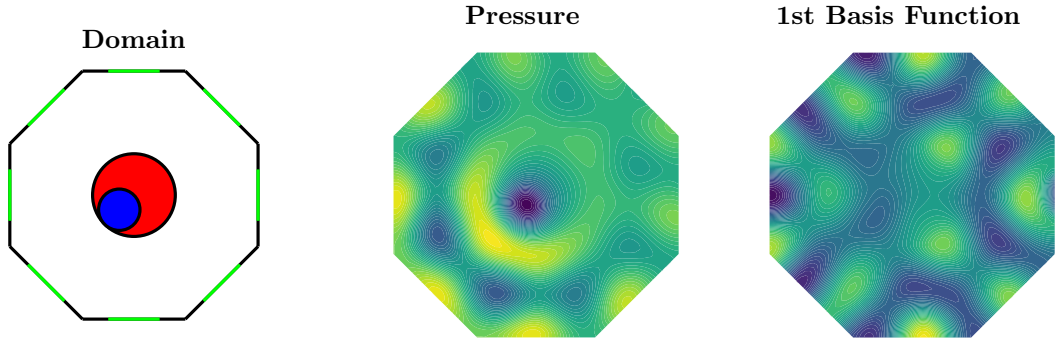

Our goal is to predict the acoustic pressure induced by a set of loud speakers using the experimental setup depicted in Figure~\ref{fig:dfat}\rev{,} which is based upon the setup in~\cite{ECStasinunas_RASchultz_MRRoss_2016a}. For a fixed angular velocity $\omega = 2\pi f$, the acoustic pressure $u$ is modeled using the (real) Helmholtz equation defined on an open regular octagon domain $D$ with apothem equal to 1.5 meters. The interior of $D$ contains a scatterer (red and blue circles) and each side of the octagon consists of an individual speaker and its cabinet; the centered green boundary segments are speaker cones which comprise $0.875$ of the total edge length and the black segments are the cabinet walls. To simplify the problem, we model the scatterer as a dense fluid and ignore the impedance of the speaker cabinet. Under these conditions, the acoustic pressure $u$ is given by
\begin{equation}\label{eq:dfat}
  \Delta u + \kappa^2 u = 0 \quad\text{in $D$},
  \qquad\qquad\frac{\partial u}{\partial n} =
    \rho_0\omega\sum_{j=1}^8 \theta_j\chi_j \quad\text{on $\partial D$}
\end{equation}
where $\kappa=\omega/c$ is the wave number, $c$ is the speed of sound, $\rho_0$ is the fluid density, $\chi_j:\partial D\to\{0,1\}$ is the characteristic function of the $j^{\text{th}}$ speaker cone (green boundary segments in Figure~\ref{fig:dfat}), and $\theta_j$ is the acoustic velocity output by the $j^{\text{th}}$ speaker for $j=1,\ldots,8$ --- in other words, the $j^{\text{th}}$ speaker cone oscillates with velocity $\theta_j\cos(\omega t)$. In this example we assume that the material in the red circle is made of aluminum for which the speed of sound is 6320 m/s and that the regions in the blue circle and exterior to the red circle are comprised of air at $20^\circ$C which has a speed of sound of 343 m/s.  In addition, we set the frequency to be $f=400$ Hz and the fluid density to be that of air at 20$^\circ$C and standard atmospheric pressure, i.e. $\rho_0=1.204$ kg/m$^3$. We discretized and solve \eqref{eq:dfat} using continuous piecewise linear finite elements.

We now use our multifidelity information fusion algorithm to predict the acoustic pressure data $u(x)$, at 5000 microphone locations $x$, using three types of experiments (information sources). 
Our aim is to predict acoustic pressure of a high-fidelity experiment, which involves activating all 8 speakers, using two lower-fidelity experiments that only activate a subset of speakers. For the high-\rev{fidelity} experiment we set the speaker amplitudes as $\theta_{3,i}=1$, $i=1,\ldots,8$ and for the low-fidelity experiments we set $\theta_{1,i}=1$, $i=3,5,7$ and $\theta_{2,i}=3$, $i=2,4,6,8$; all other speaker amplitudes are set to zero. Speakers are ordered \rev{counterclockwise} with the first speaker located on the right vertical edge of the octagon.

\revv{Given randomly selected sensor locations, we generate training data by measuring} acoustic pressure $y_k^{(i)}=u_k(x_k^{(i)})+\epsilon_k^{(i)}$ for each information source $k=1,2,3$ at random locations $x_k^{(i)}$ in the domain $D$, where the noise $\epsilon_k^{(i)}$ is normally distributed with mean zero and unit variance. \revv{We will investigate using both overlapping and non-overlapping sensor locations between the low-fidelity and high-fidelity experiments.}
In Figure~\ref{fig:dfat} (middle) we plot the true high-fidelity acoustic pressure.  Each information source
\begin{align}
  u_1(x)=\sum_{i=3,5,7} \phi_i(x)\theta_{1,i} &&   u_2(x)=\sum_{i=2,4,6,8} \phi_i(x)\theta_{2,i} && u_3(x)=\sum_{i=1}^8 \phi_i(x)\theta_{3,i}
\end{align}
is a linear sum of basis functions $\phi_i(x)$ which correspond to solving the Helmholtz equation using only one active speaker. Specifically the basis $\phi_i$ is obtained by solving
\begin{equation}
  \Delta \phi + \kappa^2 \phi = 0 \quad\text{in $D$},
  \qquad\qquad\frac{\partial \phi}{\partial n} =  
  \rho_0\omega\theta_i \quad\text{on $\partial D$}
  \label{eq:helmholtz}  
\end{equation}
The basis function $\phi_1$ is depicted in the right plot of Figure ~\ref{fig:dfat}.

Here, we augment our nonlinear least squares objective via a sparsity penalization on all of the coefficients. Specifically we use the sparse regularization~\eqref{eq:laplace_prior} objective with a single $\lambda_i=\lambda_{ij}=\lambda/2,\forall i,j$ and solve the equivalent\rev{,} but differentiable, problem
\begin{align}
  \min_{\theta,t} \sum t &+ \frac{\lambda}{2} \lVert y(x)-f(x,\theta) \rVert_2^2\\
  \text{subject to } t-\theta&\le 0\\-t-\theta&\le0
\end{align}
using the Sequential Least SQuares Programming (SLSQP) algorithm in SciPy. Here $\theta$ contains all of the parameters of the network. 
We found that the performance benefit of the multifidelity approximation is dependent on the value of the regularization parameter $\lambda$. Here we set $\lambda=1\times10^{-3}$.

\begin{figure}
  \centering
  \includegraphics[width=\textwidth]{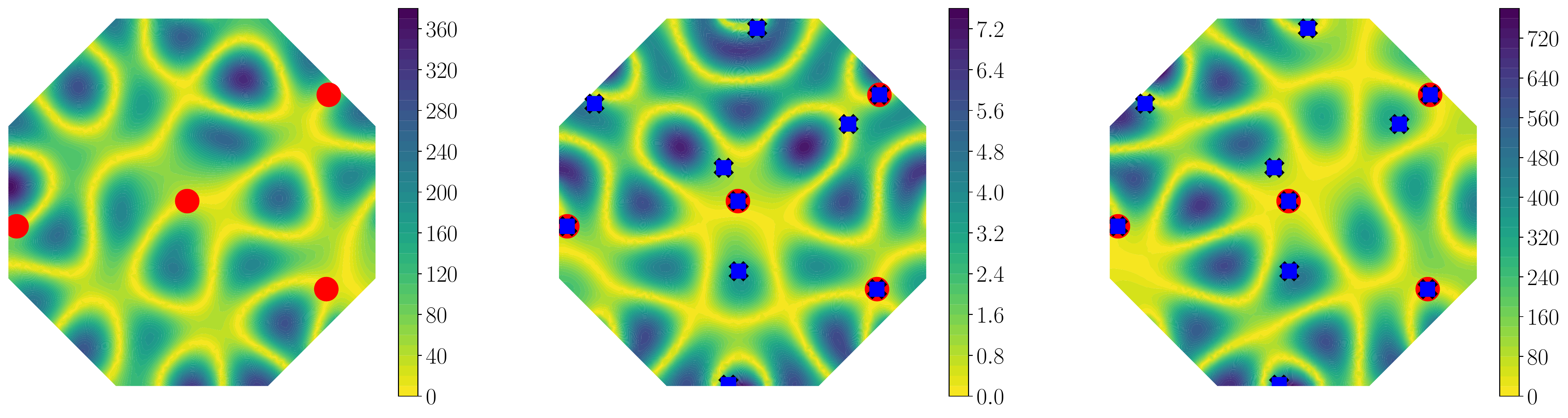}
  \includegraphics[width=\textwidth]{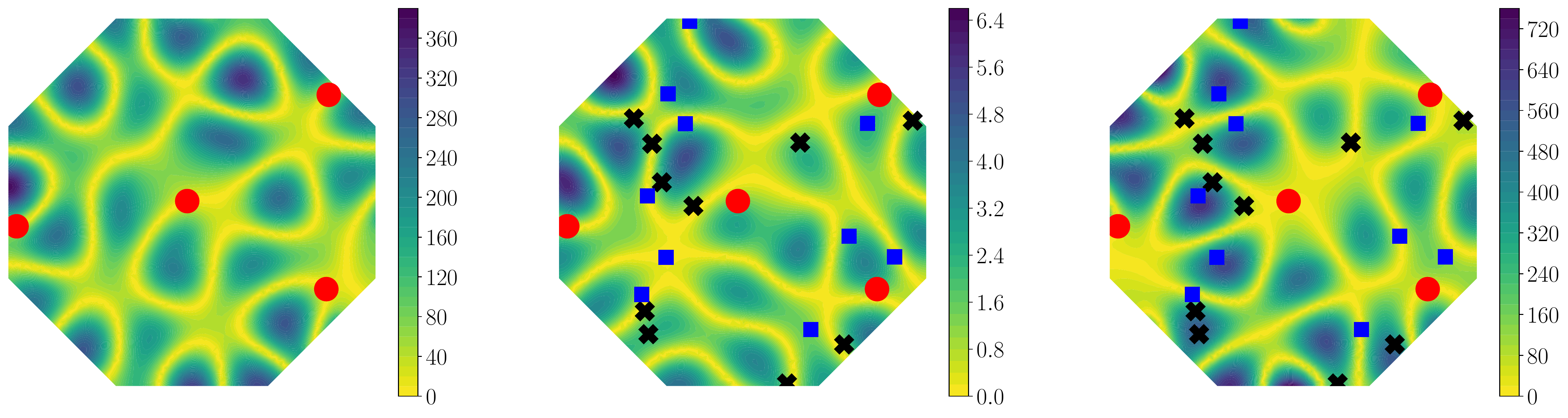}
  \caption{\revv{Direct Field Acoustic Testing. The pointwise absolute difference between the high-fidelity information source and (left), the single fidelity, (middle) the fully connected multifidelity and (right) the hierarchical ($u_2\to u_1\to u_3$) multifidelity surrogates. Circles represent the microphone locations used to extract data from the high-fidelity information source, crosses the locations used for $u_1$ and squares for $u_2$. The top row uses nested training samples the bottom row does not.}}
  \label{fig:helmholtz-error-surfaces}
\end{figure}

\revv{Next we compare the accuracy of single fidelity approximations constructed using limited high-fidelity data with multifidelity surrogates constructed with two different types of networks. Specifically we use the full graph depicted in Figure~\ref{fig:thermal-full-graph} and the hierarchical graph in Figure~\ref{fig:thermal-hier-graph}\footnote{There are two possible hierarchical orderings. We found that the errors and weight functions obtained using both orderings are almost identical and so not reported}. Furthermore, we use constant weighting functions $\rho_{ij}$.

  Figure \ref{fig:helmholtz-error-surfaces} plots the pointwise absolute differences between the true high-fidelity information source and a single-fidelity approximation and two different multifidelity approximations. The single fidelity surrogate was obtained using 4 samples of the high-fidelity source and the multifidelity approximations were obtained using an additional 10 samples of each low-fidelity source. The relative mean squared errors of the predicted acoustic pressure at the 5000 microphone locations, produced by the single-fidelity (Single), fully connected multifidelity (Full), and hierarchical multifidelity (Hier) approximations, are shown in Table~\ref{tab:dfat}. The fully connected multifidelity approximation is an order of magnitude more accurate than the single fidelity approximation and the hierarchical multifidelity approximation is less accurate than both.\footnote{The error in the full graph surrogate is dominated by the noise in the data. If noise is removed the error drops below $1\times10^{-8}$.} Note that here, unlike many existing multifidelity algorithms, we are able to train multifidelity surrogates when the high-fidelity training samples are not a subset of the low-fidelity data. Furthermore, these results are consistent regardless of whether or not the data is overlapping.}

\begin{table}[htb]
\centering
\caption{\revv{Properties of the multifidelity surrogates of the DFAT experiments. Non-nested samples are used unless otherwise stated. Dashes represent connections not present in the associated graph.}}
\label{tab:dfat}
\begin{tabular}{|c| c c c c c c|}
\hline
Surrogate Graph &   Full (nested)  &  Hier (nested)) &   Single  (nested) &   Full &  Hier &  Single \\
\hline
\hline
Error & $1.9\times10^{-3}$& $1.4\times10^{-1}$ & $7.8\times10^{-2}$& $1.1\times10^{-3}$ & $1.5\times10^{-1}$ &$7.8\times10^{-2}$\\
\hline
$\rho_{13}$ & $5.0\times10^{-1}$ & $1.0$& ---& $5.0\times10^{-1}$& $1.0$& ---\\
$\rho_{23}$ & $3.4\times10^{-1}$ & --- & --- & $3.3\times10^{-1}$& ---& ---\\
$\rho_{12}$ & $\mathbf{ 1.1\times10^{-3}}$ &  $\mathbf{ 1.1\times10^{-3}}$ &--- & $\mathbf{ 1.1\times10^{-3}}$& $\mathbf{ 1.1\times10^{-3}}$& ---\\
\hline
\end{tabular}
\end{table}

\revv{The sparse regularization we employ has a very useful effect on the learning procedure. Specifically it is able to identify unimportant connections in the multifidelity graphs. The values of the constant $\rho_{ij}$ are provided in Table~\ref{tab:dfat}. When using either the fully connected or hierarchical graph, the sparse learning algorithm identified there was no hierarchical relationship between the two low-fidelity information sources, i.e. $f_1$ does not significantly influence $f_2$, as indicated by $\rho_{12}\approx 0$ (highlighted in bold in Table~\ref{tab:dfat}). This result, suggests that sparse regularization can potentially be used to select the best graph when the true data generating graph is unknown. Future work is needed however to derive a robust algorithm for solving the non-linear $\ell^1$-minimization problem in larger graphs.}

\begin{figure}
  \centering
  \includegraphics[width=0.33\textwidth]{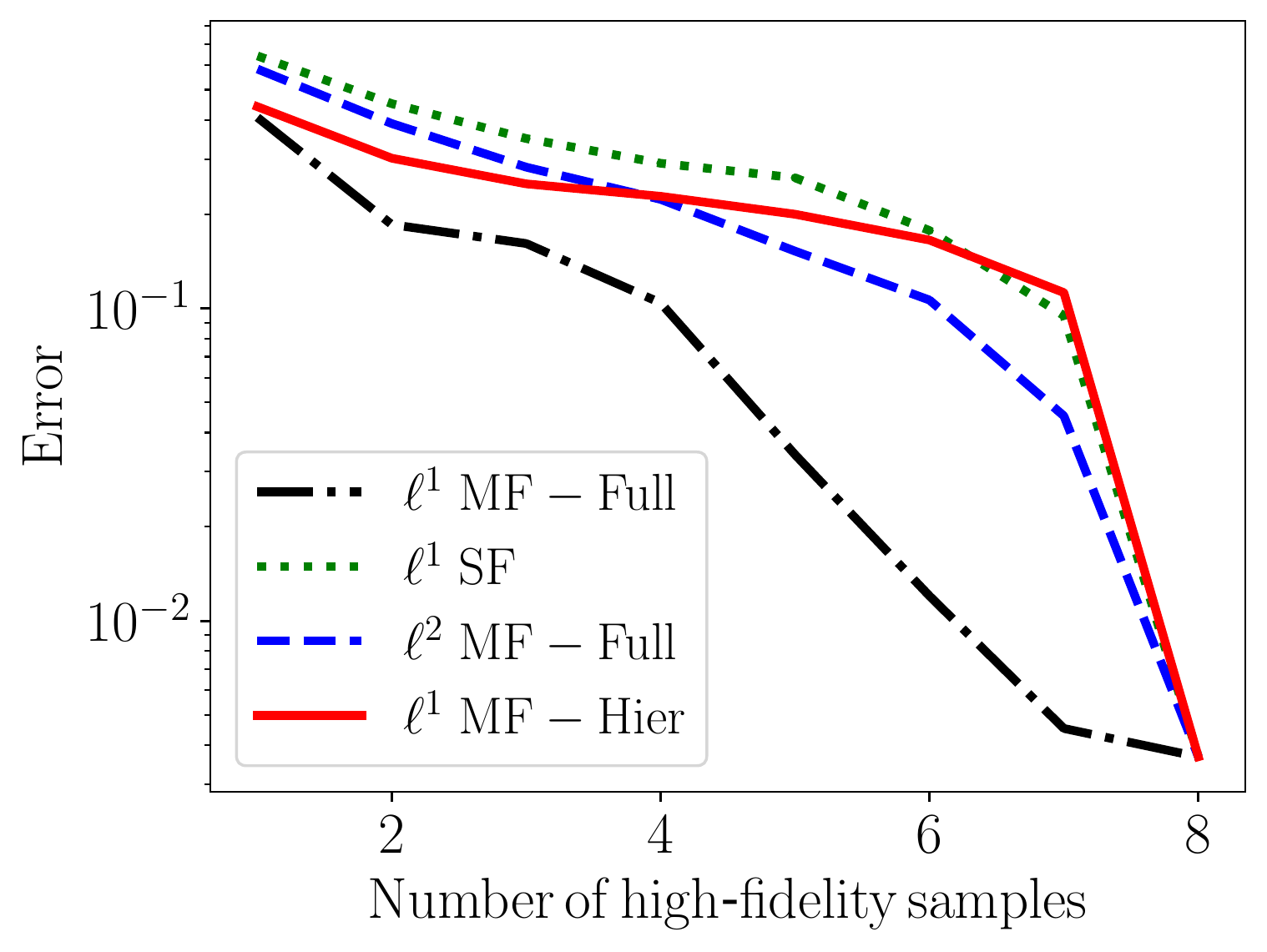}
  \includegraphics[width=0.33\textwidth]{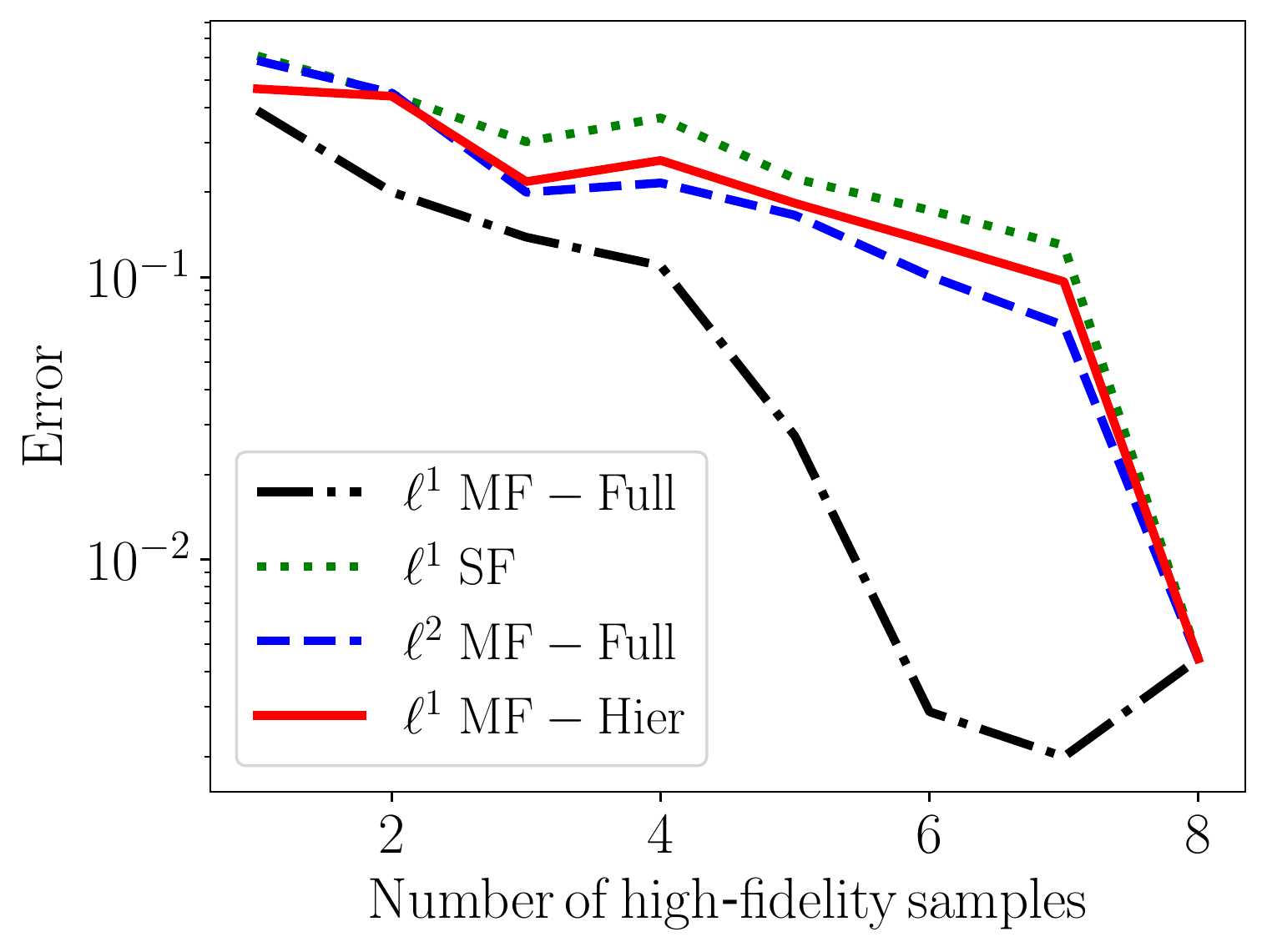}
  \caption{\revv{The relative root mean squared error at the 5000 microphone locations as the number of high-fidelity evaluations increases averaged over 10 different realizations of the training data. (Left) nested samples and (right) non-nested samples. The number of evaluations of both the low-fidelity information sources is fixed at 10. The label $\ell^1$ refers to an approximation built via a Laplace prior regularization\eqref{eq:laplace_prior}, and $\ell^2$ refers to only minimizing the negative log likelihood~\eqref{eq:graph-neg-log-like}. MF refers to a multifidelity approximation and SF a single fidelity approximation.}}
  \label{fig:helmholtz-error-convergence}
\end{figure}

\revv{Next we discuss the impact of  training data on the accuracy of the surrogates by repeating these experiments over ten realizations of the data. In Figure \ref{fig:helmholtz-error-convergence} (right) we plot the average root mean \rev{squared} error}
\begin{align*}\frac{\lVert u_3-\hat{u}_3\rVert_{\ell^2}}{\lVert u_3\rVert_{\ell^2}}&&\lVert g\rVert_{\ell^2}=\sum_{i=1}^{5000}g(x_k^{(i)})\end{align*}
in the multifidelity approximation $\hat{f}_3$ as the number of high-fidelity samples increases while number of low-fidelity experiments is fixed at 10. \revv{The multifidelity approximation based upon the fully connected graph $\ell^1\; \mathrm{MF}-\mathrm{Full}$ that enforces sparsity is much more accurate than the other approximation types. However, removing the sparsity promoting regularization degrades the accuracy of the fully connected surrogate $\ell^2\; \mathrm{MF}-\mathrm{Full}$. The single fidelity approximation $\ell^1\; \mathrm{SF}$ and the hierarchical multi-fidelity surrogate $\ell^1\; \mathrm{MF}-\mathrm{Hier}$ that enforce sparsity consistently have the largest error. \revv{All methods reach the same accuracy when 8 high-fidelity evaluations are used. At this point the noise in the data dominates the surrogate error.\footnote{\revv{The absolute standard deviation of the noise is 1, but the relative standard deviation, normalized by $\lVert u_3\rVert_2$ (the same factor used to normalize the relative error) is $6.3\times10^{-4}$. This implies that 3 standard deviations of relative noise is approximately $2\times10^{-3}$, which roughly corresponds to the minimum error in Figure~\ref{fig:helmholtz-error-convergence}.}} As more high-fidelity evaluations ($>8$) are used, the error in the surrogates produced by all methods will converge at the same rate. Additional evaluations only decrease the impact of noise.}

  In summary, sparsity and low-fidelity data is needed to produce an accurate prediction with limited high-fidelity data. When enough high-fidelity data is obtained all approximations have similar error and these conclusions do not seem to be significantly impacted by the use of nested or non-nested training data.}

\section{Conclusion}

In this paper we have developed, analyzed, and numerically demonstrated a multifidelity information fusion approach that enables extremely flexible modeling of known relationships amongst information sources. We have shown that this approach can yield significantly more accurate surrogate models than the predominant hierarchical approaches found in the literature. In particular, we have shown that while hierarchical approaches can be shown to have equivalent expressivity as more general models, they make use of data less efficiently. Indeed\rev{,} for the low-data settings, exploiting more complex, but often more natural, structure can become extremely beneficial.

We envision that the proposed approach will increase the applicability of general multi-level and multifidelity approaches in uncertainty quantification and data-driven learning to areas with less traditional relationships between data sources (e.g., not arising from a hierarchy of discretizations or reduced order models). Future work will require both data-driven discovery of optimal network structures as well as physics and numerics driven derivation of optimal network structures in different application areas. \revv{The results presented in this paper suggest that sparse regularization can be used to select the best graph when the true data generating graph is unknown. However, further work is needed to determine the veracity of this hypothesis.}

\section*{Acknowledgments}
The authors were supported by the Laboratory Directed Research Development (LDRD) program at Sandia National Laboratories. Sandia National Laboratories is a multi-mission laboratory managed and operated by National Technology and Engineering Solutions of Sandia, LLC., a wholly owned subsidiary of Honeywell International, Inc., for the U.S. Department of Energy's National Nuclear Security Administration under contract DE-NA-0003525. The views expressed in the article do not necessarily represent the views of the U.S. Department of Energy or the United States Government.